\documentclass[sigconf]{acmart}

\renewcommand\footnotetextcopyrightpermission[1]{} % removes footnote with conference information in first column
\usepackage{fancyhdr}
\usepackage{balance}
\usepackage{multirow}

\usepackage{amsmath,amssymb,amsfonts}
\usepackage{bm}
\usepackage{subfigure}
\usepackage{comment}
\usepackage{adjustbox}

\AtBeginDocument{%
  \providecommand\BibTeX{{%
    \normalfont B\kern-0.5em{\scshape i\kern-0.25em b}\kern-0.8em\TeX}}}

\makeatletter % For subfigure omit (a)(b)(c)
\renewcommand{\@thesubfigure}{\hskip\subfiglabelskip}
\makeatother

\begin{document}


\title{AFD-Net: Adaptive Fully-Dual Network for Few-Shot Object Detection}

%%
% The "author" command and its associated commands are used to define
% the authors and their affiliations.
% Of note is the shared affiliation of the first two authors, and the
% "authornote" and "authornotemark" commands
% used to denote shared contribution to the research.
\author{Longyao Liu, Bo Ma, Yulin Zhang, Xin Yi and Haozhi Li}
\affiliation{\institution{Beijing Institute of Technology}}
\email{roel\_liu@bit.edu.cn}

\begin{abstract}
	Few-shot object detection (FSOD) aims at learning a detector that can fast adapt to previously unseen objects with scarce annotated examples, which is challenging and demanding. Existing methods solve this problem by performing subtasks of classification and localization utilizing a shared component (\emph{e.g.}, RoI head) in the detector, yet few of them take the distinct preferences of two subtasks towards feature embedding into consideration. In this paper, we carefully analyze the characteristics of FSOD, and present that a general few-shot detector should consider the explicit decomposition of two subtasks, as well as leveraging information from both of them to enhance feature representations. To the end, we propose a simple yet effective Adaptive Fully-Dual Network (AFD-Net). Specifically, we extend Faster R-CNN by introducing Dual Query Encoder and Dual Attention Generator for separate feature extraction, and Dual Aggregator for separate model reweighting. Spontaneously, separate state estimation is achieved by the R-CNN detector. Besides, for the acquisition of enhanced feature representations, we further introduce Adaptive Fusion Mechanism to adaptively perform feature fusion in different subtasks. Extensive experiments on PASCAL VOC and MS COCO in various settings show that, our method achieves new state-of-the-art performance by a large margin, demonstrating its effectiveness and generalization ability.
\end{abstract}

\begin{CCSXML}
	<ccs2012>
	<concept>
	<concept_id>10010147.10010178.10010224.10010245.10010250</concept_id>
	<concept_desc>Computing methodologies~Object detection</concept_desc>
	<concept_significance>500</concept_significance>
	</concept>
	</ccs2012>
\end{CCSXML}

\ccsdesc[500]{Computing methodologies~Object detection}

\keywords{Few-Shot Object Detection, Meta-Learning, Task Decomposition, Adaptive Feature Fusion}

\maketitle
\section{Introduction}
\label{sec:introduction}
Recent years have witnessed impressive advances of convolutional neural networks (CNNs)~\cite{he2016deep,dai2017deformable,huang2017densely} in object detection~\cite{ren2015faster,redmon2017yolo9000,dai2016r,girshick2014rich,lin2017feature,redmon2016you,cai2018cascade,girshick2015fast}, due to the availability of large-scale benchmarks with accurate annotations~\cite{russakovsky2015imagenet,everingham2010pascal,lin2014microsoft}. However, training general object detection models from scratch typically requires rich labeled data, which is extremely expensive to obtain or even hard to collect, such as endangered animals or certain medical data. Thus, detectors significantly suffer a performance drop when training examples are inadequate~\cite{zhang2016understanding}. On the contrary, humans exhibit a strong ability to address this issue: even a child can easily learn to recognize novel characteristics from only a few instances~\cite{samuelson2005they}.

%------------------------------------------------------------------------
\begin{figure}[!t]
	\centering
	\includegraphics[width=0.95\linewidth]{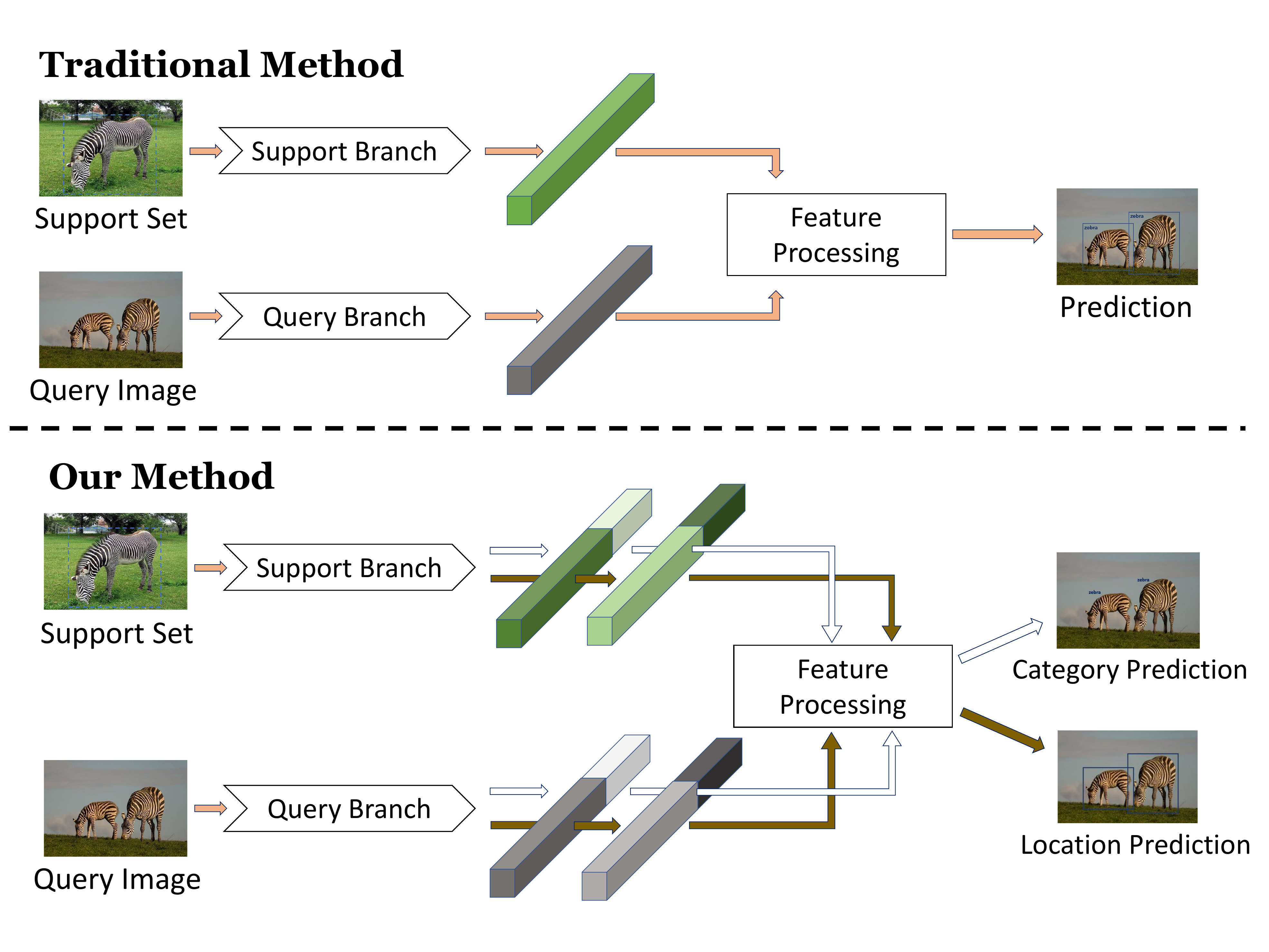}
	\vspace{-0.3cm}
	\caption{\small Comparison of our framework and traditional methods. Traditional methods use shared encoded features to perform the estimation of object categories and locations, while our Adaptive Fully-Dual Network decomposes the tasks of classification (white arrow) and localization (brown arrow). Besides, feature fusion mechanism (concatenating encoded features with different colors) is introduced for enhancing feature representations.}
	\label{fig:1}
\end{figure}
%------------------------------------------------------------------------

This triggers recent researches on few-shot learning (FSL). It is considered promising to enhance the  generalization ability of deep networks from limited training examples~\cite{lake2015human,vinyals2016matching,finn2017model,nichol2018first,andrychowicz2016learning}. Concretely, FSL aims at recognizing instances from novel classes given only a few annotated data per category in the inference stage, with the availability of abundant labeled training samples from base classes. Most researches in few-shot learning community focus on image classification~\cite{koch2015siamese,vinyals2016matching,snell2017prototypical,sung2018learning,gidaris2018dynamic,qiao2018few}, while far less progress has been made in the field of object detection~\cite{kang2019few,yan2019meta,Xiao2020FSDetView}, which is generally considered much more challenging due to the existence of an additional subtask, \emph{i.e.}, few-shot localization, besides the subtask of object recognition.

A trend to solve few-shot object detection is to conjoin a reweighting module with a base object detector, \emph{e.g.}, Faster R-CNN~\cite{ren2015faster}. Concretely, these few-shot detectors~\cite{yan2019meta,Xiao2020FSDetView} introduce an additional branch to extract discriminative features of support set from novel classes, and then use these features to reweight RoI (Region-of-Interest) head, which is shared for subsequent estimation of object categories and locations. However, classification and localization are two fundamentally different subtasks in few-shot object detection. The former subtask focuses on providing a coarse location of the object via classification, while the latter aims at estimating an accurate object state by a refined bounding box~\cite{xu2020siamfc++,wu2020rethinking}. This leads to distinct preferences towards feature representation of these two subtasks. In other words, generated features for classification are probably not suitable for bounding box regression, and using the same RoI representations in two subtasks is suboptimal. Owing to the explicit task decomposition, in the process of a specific subtask, the obtained features for state estimation can be enriched by the information from the unfocused subtask. Therefore, it is crucial to exploit an efficient feature fusion mechanism between tasks of classification and location regression.

Motivated by the abovementioned analysis, we present that a general few-shot detector should: 1) decouple the process of classification from that of localization, involving feature representations, model reweighting and state estimation; 2) introduce efficient feature fusion mechanism between two subtasks into the encoder of few-shot detector, to enhance feature representations for both query image and support set, as shown in Fig.~\ref{fig:1}.

\begin{figure*}[!t]
	\vspace{-1.5cm}
	\centering
	\includegraphics[width=1.0\linewidth]{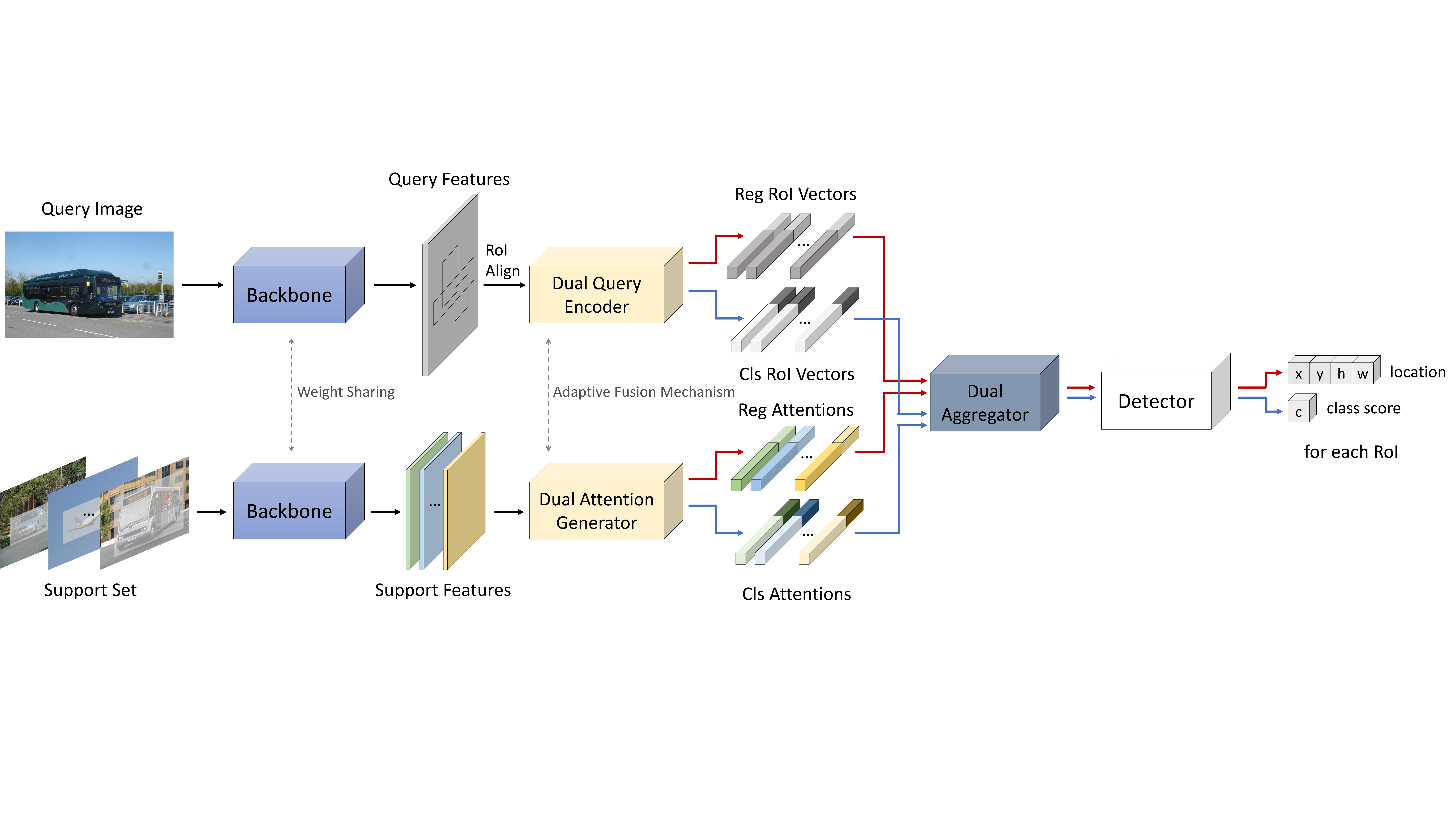}
	\vspace{-2.8cm}
	\caption{\small The pipeline of Adaptive Fully-Dual Network for few-shot object detection. It consists of three key components: 1) the top query branch receives a query image and generates its RoI vectors; 2) the bottom support branch encodes support set into its class-attentive vectors; 3) two groups of features from two branches are aggregated for subsequent estimation of object categories and locations with a R-CNN detector. We further introduce three modules, \emph{i.e.}, Dual Query Encoder, Dual Attention Generator and Dual Aggregator in three components, along with Adaptive Fusion Mechanism for enhancing feature representations. All these three modules are in a dual architecture, where the specific subtask is performed in the assigned path. Black line represents shared path for two subtasks, {\color{red} red} and {\color{blue} blue} lines represent subtasks of bounding box regression and category classification, respectively.}
	\label{fig:2}
\end{figure*}

In this paper, we propose a novel and intuitive framework for few-shot object detection, namely \textit{Adaptive Fully-Dual Network (AFD-Net)}, as illustrated in Fig.\ref{fig:2}. Specifically, we extend Faster R-CNN by introducing three modules, \emph{i.e.}, Dual Query Encoder (DQE), Dual Attention Generator (DAG), and Dual Aggregator (DA), for separate process of two subtasks in various key stages of FSOD. Besides, Adaptive Fusion Mechanism (AFM), guiding the design of DQE and DAG, is introduced for the acquisition of enhanced feature representations. Query features generated by backbone network are encoded by DQE into two groups of RoI vectors. These vectors contain meta information generalizable to detect novel objects within RoIs, and each group is utilized for a specific subtask. In parallel with DQE, DAG takes support images as input and encodes them into class-attentive vectors in the same scheme. DA, consisting of two aggregators for two subtasks, performs separate model reweighting. Each RoI vector from query image and each class-attentive vector from support set assigned the same subtask are aggregated in the corresponding aggregator. In this way, some meta features of query image informative for detecting novel objects would be activated. Finally, the aggregated features are fed into the R-CNN detector to estimate the object location and category, respectively, achieving separate state estimation, spontaneously.

Extensive experiments on two public datasets, \emph{i.e.}, PASCAL VOC and MS COCO, in various settings, show that despite its simplicity, our method outperforms state-of-the-art approaches by a large margin, demonstrating its effectiveness and generalization ability.

In summary, the main contributions of this paper are three-fold:
\begin{itemize}
	\item We propose a simple yet effective framework for few-shot object detection. To the best of our knowledge, we are among the first to solve FSOD by decomposing the process of classification and localization.
	
	\item We further introduce three modules, \emph{i.e.}, Dual Query Encoder, Dual Attention Generator and Dual Aggregator, to perform decomposition in multiple components of our network, along with Adaptive Fusion Mechanism for enhancing feature representations.
	
	\item The experimental results demenstrate that our proposed method achieves new state-of-the-art performance on multiple benchmarks.
\end{itemize}

\section{Related Work}
\label{sec:RelatedWork}
{\bfseries General Object Detection.}
Recent object detectors based on deep CNNs can be mainly divided into two steams, \emph{i.e.}, one-stage detectors and two-stage ones. R-CNN series~\cite{girshick2014rich,girshick2015fast,ren2015faster,he2015spatial,dai2016r,lin2017feature} belong to the first category, which generate region proposals~\cite{ren2015faster} of potential objects in the first stage, and then  perform category and bounding box estimation at the proposal-level. On the contrary, YOLO~\cite{redmon2016you} and the variants~\cite{redmon2017yolo9000,liu2016ssd,redmon2018yolov3,bochkovskiy2020yolov4} dominate the one-stage steam. These methods use a single CNN to predict categories and locations of the objects directly, without explicitly generating proposals. These two branches of general object detectors unanimously depend on a huge amount of data with elaborate bounding box annotations. When training samples are limited, they struggle heavily.

{\bfseries Few-shot Learning.}
Few-shot learning refers to learning to learn general knowledge that can be easily transferred to new tasks with only a handful of annotated examples~\cite{lake2015human,vinyals2016matching,andrychowicz2016learning}. Few-shot classification has recently been widely investigated as a representative task of few-shot learning. Generally, solutions to this problem involve two groups: meta-learning based and fine-tuning based methods. The former can be further categorized into: 1) metric-learning based methods~\cite{koch2015siamese,snell2017prototypical,sung2018learning} that focus on the similarity of input images in the embedding space; 2) optimization-based ones~\cite{ravi2016optimization,finn2017model,nichol2018first}, where a meta-learner is designed to simulate the optimization process for the fast adaptation to novel classes with limited samples; 3) model-based methods~\cite{bertinetto2016learning,gidaris2018dynamic,wang2019tafe} that aim to estimate the network parameters for novel tasks with the help of a learned predictor. Fine-tuning based approaches~\cite{chen2019closer,chen2020new} demonstrate that simple fine-tuning techniques are crucial and effective towards few-shot learning.

{\bfseries Few-shot Object Detection.}
While substantial progress has been made in few-shot classification, the problem of few-shot object detection is relatively unsolved, since object localization should be additionally processed in the scenario where distracting irrelevant objects may exist or even there is no object within an image. LSTD~\cite{chen2018lstd} applies transfer learning technique in object detection with limited examples and  demonstrates its effectiveness. TFA~\cite{wang2020frustratingly} proposes that only fine-tuning the last layer of existing detectors is crucial and effective despite its simplicity. Metric learning widely explored in few-shot classification can be extended to few-shot object detection. For example, RepMet~\cite{karlinsky2019repmet} scores the pair-wise similarity between embedded features of input query and support images. \cite{fan2020few} incorporates attention mechanism into RPN (Region Proposal Network) and explicitly defines a multi-relation detector to suppress detection of the background. Recently, methods based on meta-learning have been a popular trend. FSRW~\cite{kang2019few} attaches a reweighting module with YOLOv2~\cite{redmon2017yolo9000} to adjust the full input image features using class-attentive vectors of support images, while Meta R-CNN~\cite{yan2019meta} applies reweighting on RoI features based on Faster R-CNN~\cite{ren2015faster}. MetaDet~\cite{wang2019meta} presents a meta-model to predict the parameters of category-agnostic and category-specific components in a detector, separately. FSDetView~\cite{Xiao2020FSDetView} applys a slightly more complicated feature aggregation scheme to further improve the detection performance and evaluation reliability.

\section{Methodology}
\label{sec:Methodology}
\subsection{Problem Setup}\label{M_A}
As in previous work~\cite{kang2019few,yan2019meta,Xiao2020FSDetView}, we adopt the following few-shot object detection settings. In the training phase, we are provided with two training sets, \emph{i.e.}, a base set $D_{\text{\textit{base}}}$ from base classes $C_{\text {\textit{base}}}$ with abundant instances and a novel set $D_{\text {\textit{novel}}}$ from novel classes $C_{\text {\textit{novel}}}$ with only a few samples per category, where $C_{\text{\textit{base}}}$ and $C_{\text {\textit{novel}}}$ are non-overlapping, \emph{i.e.}, $C_{\text {\textit{base}}} \cap C_{\text {\textit{novel}}}=\varnothing$. For each sample $(x, y) \in D_{\text {\textit{base}}} \cup D_{\text {\textit{novel}}}$, $x=\left\{\textit{obj}_{i}\right\}_{i=1}^{N}$ is an image containing $N$ objects, and $y=\left\{\left(\textit{cls}_{i}, \textit{box}_{i}\right)\right\}_{i=1}^{N}$ denotes $N$ categories $\left\{\textit{cls}_{i}\right\}_{i=1}^{N}$ with each $\textit{cls}_{i} \in C_{\text {\textit{base}}} \cup C_{\text {\textit{novel}}}$ along with $N$ structured annotations $\left\{\textit{box}_{i}\right\}_{i=1}^{N}$ of the $N$ objects in the image $x$. The aim of the few-shot object detector is to classify and locate objects from both novel and base classes in an image with only $K$ (usually less than 10) available instances per class in the phase of inference, with the help of transferable knowledge learned from abundant examples from base classes.

\subsection{Adaptive Fully-Dual Network}\label{M_B}

{\bfseries Pipeline.}
Our proposed Adaptive Fully-Dual Network decomposes the process of few-shot classification and localization, as well as enhancing feature representations. We adopt the widely used two-stage detector Faster R-CNN~\cite{ren2015faster} as the base model, which first generates RoIs of potential objects, and then performs state estimation using a single RoI head. The pipeline of our proposed network is demonstrated in Fig.~\ref{fig:2}. Concretely, we extend Faster R-CNN to a dual Siamese architecture, where a query image is encoded by the top Faster R-CNN branch, and the bottom branch is designed for support set.  Each branch further contains two paths, and feature extraction for the specific subtask is performed in the corresponding path. Generated features from two branches are then aggregated, achieving the reweighting of Faster R-CNN for the subsequent detection of novel instances. We introduce Dual Query Encoder (DQE) and Dual Attention Generator (DAG) for separate feature representations, along with Dual Aggregator (DA) for separate model reweighting, in our proposed architecture. Furthermore, we present Adaptive Fusion Mechanism (AFM) to efficiently encode both the query image and support set.

{\bfseries Dual Feature Representations.}
Adaptive Fully-Dual Network consists of two branches, \emph{i.e.}, the top Faster R-CNN branch for processing the query image and the bottom support branch for support set, as shown in Fig. \ref{fig:2}. The Faster R-CNN branch aims to learn meta features generalizable to detect novel objects within the input query image. The support branch encodes the input support set into discriminative vectors, which will be subsequently utilized to adjust the contribution of meta features generated from Faster R-CNN branch. In this way, meta features informative for detecting novel objects would be activated. Therefore, these support vectors, namely class-attentive vectors, can be seen as the attention coefficients of meta features.

In two-stage object detectors, for an input image $Q$, RPN is applied to generate $n$ class-agnostic RoIs. These RoIs are then embeded by the RoI head into RoI vectors for subsequent state estimation. Instead of generating a single group of RoI vectors $\left\{r_{i}\right\}_{i=1}^{n}$ for the input query image $Q$ by a single RoI head in previous work~\cite{yan2019meta,Xiao2020FSDetView}, we propose Dual Query Encoder $\mathcal{E}$ in Faster R-CNN branch, to obtain two groups of $m$-dimensional RoI vectors $\left\{r_{i}^{cls},r_{i}^{reg}\right\}_{i=1}^{n} \in \mathbb{R}^{2n\times{m}}$ containing meta information for subtasks of classification and bounding box regression respectively, guided by our observation that these two subtasks in few-shot detection should be treated separately. This procedure is formulated as below:
\begin{equation} \label{eq:1}
	\left\{r_{i}^{cls},r_{i}^{reg}\right\}_{i=1}^{n}=\mathcal{E}(\mathcal{R}(\mathcal{B}(Q)))
\end{equation}
where $\mathcal{B}$ denotes the backbone network of Faster R-CNN, and $\mathcal{R}$ denotes the RoIAlign operation.

As for the support branch, it takes support images as input and generates their class-attentive vectors for subsequently reweighting meta features from Faster R-CNN branch. To effectively capture the support information, the input support image $S$ is concatenated with a structured binary mask $M$ indicating the bounding box annotation of the target object to detect~\cite{kang2019few}. Supposing there are $s$ categories in the support set. Support features of input support image concatenated with its associated mask $[S_{j},M_{j}]$ from class $j, j=1,2,\dots,s$, are generated from backbone network $\mathcal{B}$ sharing weights with Faster R-CNN branch. Then our proposed Dual Attention Generator $\mathcal{G}$ embeds them into two $m$-dimensional class-attentive vectors $\left\{a_{j}^{cls},a_{j}^{reg}\right\} \in \mathbb{R}^{2\times{m}}$ for category classification and bounding box regression, respectively:
\begin{equation} \label{eq:2}
	\left\{a_{j}^{cls},a_{j}^{reg}\right\}=\mathcal{G}(\mathcal{B}[S_{j},M_{j}])
\end{equation}

Each RoI vector of query image and each class-attentive vector of support set assigned the same subtask will be aggregated by our proposed Dual Aggregator $\mathcal{D}$ before being fed into R-CNN detector for state estimation. We will discuss the details next.

In this paper, the input query image $Q$ and support image $S$ are resized into $224\times224$. The backbone network $\mathcal{B}$ is the first four ResNet blocks, and the input dimension of support branch is $224\times224\times4$. The size of each query RoI is $7\times7\times1024$, and that of each support feature is $14\times14\times1024$. All of the RoI vectors and class-attentive vectors are $3072$-dimensional.

%--------------------------------------------------------------------------
\begin{figure}[t] %\centering
	\vspace{-0.2cm}
	\centering
	\includegraphics[width=1.0\linewidth]{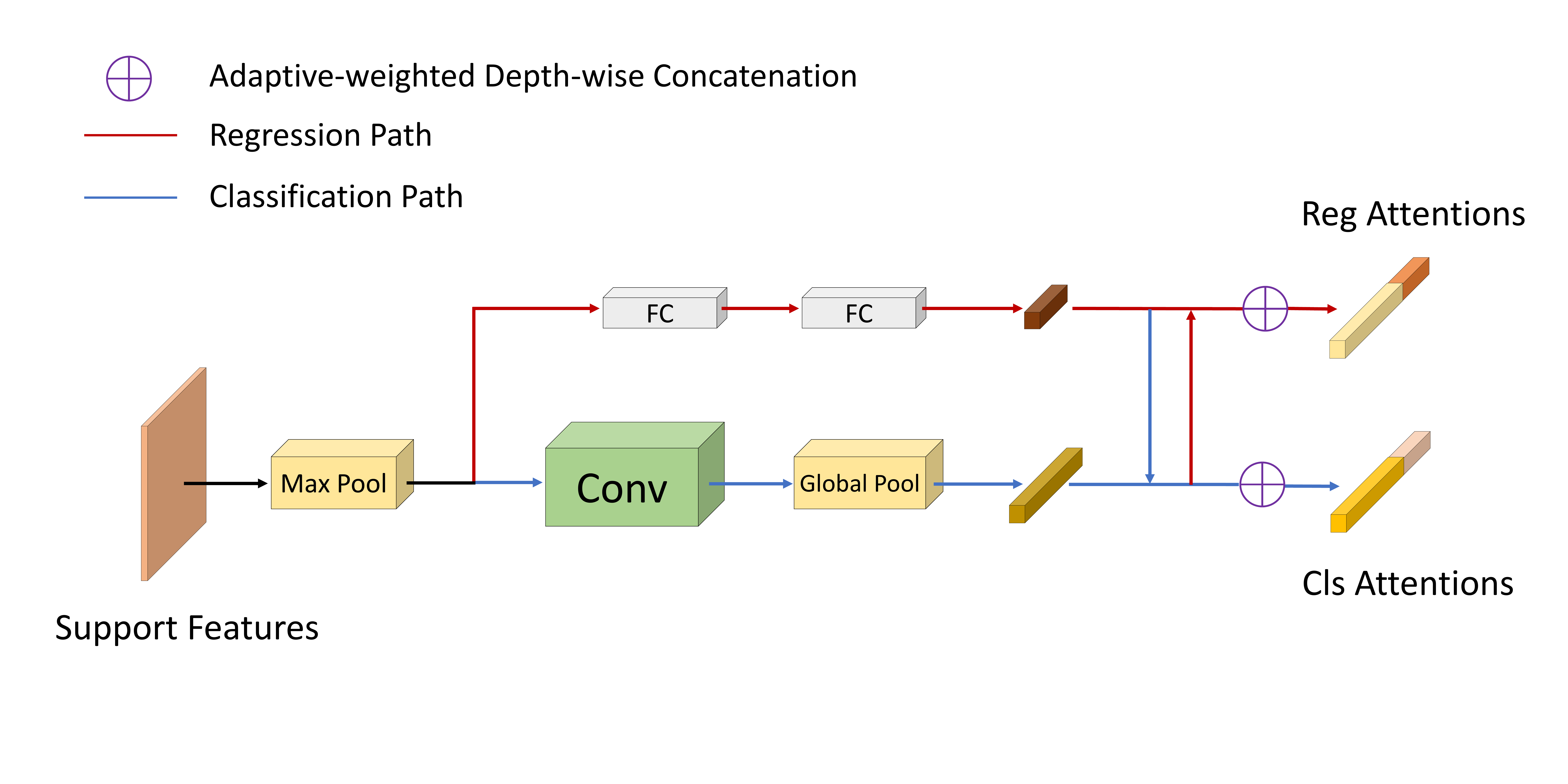}
	\vspace{-1.1cm}
	\caption{\small Illustration of Adaptive Fusion Mechanism.}
	\vspace{-0.3cm}
	\label{fig:3}
\end{figure}
%--------------------------------------------------------------------------

{\bfseries Adaptive Fusion Mechanism.}
To enhance feature representations for both the query and support set, we propose Adaptive Fusion Mechanism. Its key philosophy is performing feature extraction for two subtasks in distinct manners, as well as using the information from the unfocused subtask for further enhancing feature representations. Adaptive Fusion Mechanism guides the design of Dual Query Encoder and Dual Attention Generator. Actually these two modules share weights except for the additional Max Pool layer in Dual Attention Generator, and we take this module as an example to elaborate this mechanism, as shown in Fig. \ref{fig:3}.

Dual Attention Generator firstly applies a Max Pool layer to ensure the consistency in the size of input feature with Dual Query Encoder. Then it involves two parallel branches and each branch encodes input support images into class-attentive vectors for the specific subtask $t$, \emph{i.e.}, $cls$ for category classification and $reg$ for bounding box regression. We adopt a two-layer fully-connected (\textit{2-fc}) encoder $\mathcal{G}_{fc}$ in regression branch and a convolution (\textit{conv}) encoder $\mathcal{G}_{conv}$ in classification branch. Output task-specific class-attentive vectors from two branches are the concatenation of weighted \textit{fc} features and \textit{conv} features. The weights can be adaptively adjusted according to the specific subtask, and their values indicate the contributions to the integrated task-specific outputs. Therefore, Eq.~(\ref{eq:2}) can be rewritten as:
\begin{equation} \label{eq:3}
	\left\{a_{j}^{t}\right\}=\left[\lambda_\textit{conv}^{t}\mathcal{G}_{\textit{conv}}\mathcal{B}([S_{j},M_{j}]),\lambda_\textit{fc}^{t}\mathcal{G}_{\textit{fc}}\mathcal{B}([S_{j},M_{j}])\right], t \in\{\textit{cls},\textit{reg}\}
\end{equation}
where $\lambda_{\text {\textit{conv}}}^{t}$ and $\lambda_{\textit{fc}}^{t}$ denote the learnable weights of corresponding task-specific class-attentive vectors and are initialized to 1, $[\cdot,\cdot]$ denotes the depth-wise concatenation operation.

In this paper, the kernel size of Max Pool layer is $2\times2$. The \textit{conv} encoder $\mathcal{G}_{conv}$ is the last ResNet block, and the Global Pool layer encodes the output \textit{conv} features into a $2048$-dimensional vector. Two layers of \textit{fc} encoder $\mathcal{G}_{fc}$ are both $1024$-dimensional, thus the output vector of $\mathcal{G}_{fc}$ is $1024$-dimensional. Output task-specific class-attentive vectors are $3072$-dimensional.

%--------------------------------------------------------------------------
\begin{figure}[t] %\centering
	\vspace{-0.3cm}
	\begin{center}
		\includegraphics[width=1.0\linewidth]{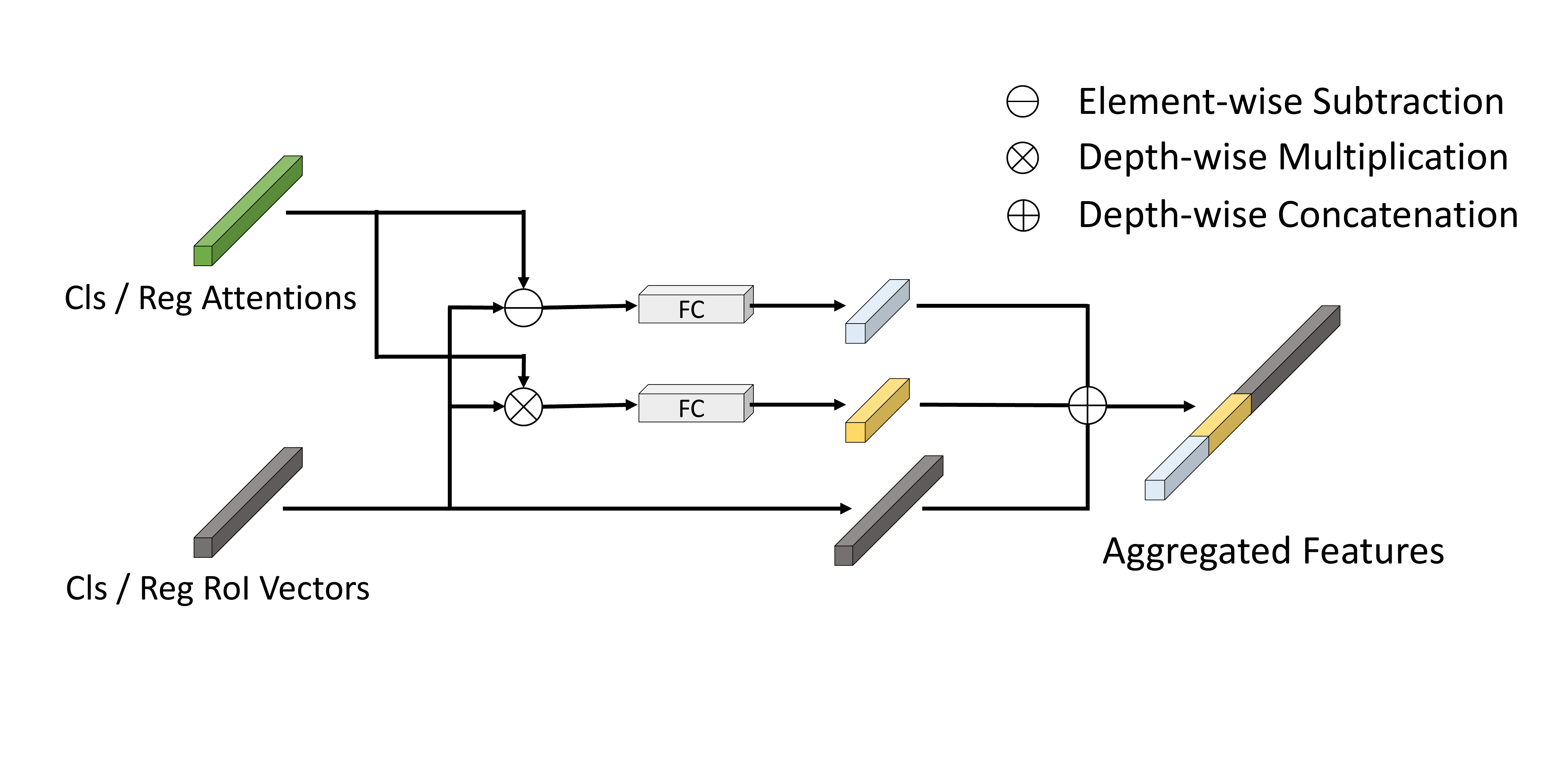}
	\end{center}
	\vspace{-0.7cm}
	\caption{\small Illustration of feature aggregation scheme.}
	\vspace{-0.5cm}
	\label{fig:4}
\end{figure}
%-------------------------------------------------------------------------

{\bfseries Dual Feature Aggregation.}
The class-attentive vectors are responsible for reweighting RoI vectors and activating those  that encode potential novel objects. The aggregated RoI vectors are then fed into the R-CNN detector for state estimation. In this paper, besides separate feature representations for two subtasks, we decompose the process of feature aggregation to realize separate model reweighting, leading to our Dual Aggregator $\mathcal{A}$. It consists of a classification aggregator and a bounding box regression aggregator in parallel. In each aggregator, each RoI vector of query image is aggregated with each class-attentive vector of support features following the scheme introduced in~\cite{Xiao2020FSDetView}, as shown in Fig.~\ref{fig:4}, which can be formulated as:
\begin{comment}
\begin{equation}
\begin{aligned}
\left\{f_{k}^{t}\right\}_{k=1}^{n \times s} &=\mathcal{A}^{t}\left(\left\{r_{i}^{t}\right\}_{i=1}^{n},\left\{a_{j}^{t}\right\}_{j=1}^{s}\right) \\
&=\left\{\left[r_{i}^{t} \otimes a_{j}^{t}, r_{i}^{t}-a_{j}^{t}, r_{i}^{t}\right]_{k}^{t}\right\}_{k=1}^{n \times s}, t \in\{\textit{cls}, \textit{reg}\} \\
%&\text {for each } i=1, \cdots, n, j=1, \cdots, m \\
&\text {\textit{for each} } i \in\{1,2, \cdots, n\}, j \in\{1,2, \cdots, s\}
\end{aligned}
\end{equation}
\end{comment}
\begin{equation}
	\begin{aligned}
		r_{i,j}^{t} &=\mathcal{A}\left(r_{i}^{t},a_{j}^{t}\right) \\
		&=\left[f_m(r_{i}^{t} \otimes a_{j}^{t}), f_s(r_{i}^{t}-a_{j}^{t}), r_{i}^{t}\right], t \in\{\textit{cls}, \textit{reg}\} \\
		%&\text {for each } i=1, \cdots, n, j=1, \cdots, m \\
		&\text {\textit{for each} } i \in\{1,2, \cdots, n\}, j \in\{1,2, \cdots, s\}
	\end{aligned}
\end{equation}
where $r_{i,j}^{t}$ denotes the $i$-th RoI vector reweighted by the $j$-th class-attentive vector from the aggregator assigned the specific subtask $t$, and $\otimes$ denotes the depth-wise multiplication implemented through $1\times1$ depth-wise convolution. $f_m$ and $f_s$ are FC layers in Fig.~\ref{fig:4}, and they are both $1536$-dimensional. Therefore, the output aggregated features are $6144$-dimensional.

{\bfseries Dual State Estimation.}
As the input features of R-CNN detector are divided into two groups $r_{i,j}^{cls}$ and $r_{i,j}^{reg}$ where $i=1,2,\dots,n$, $j=1,2,\dots,s$, the state estimation in our framework is decomposed as well. Specifically, the aggregated RoI vector $r_{i,j}^{cls}$ is utilized for predicting the probability of the $i$-th RoI containing an object from novel class $j$. Therefore, the classifier will produce $s$ outcomes for each RoI, and the category $h$ with the highest confidence score will be assigned. Then the associated RoI vector $r_{i,h}^{reg}$ will be used for location estimation of the $i$-th RoI by the regressor. If the highest confidence score is lower than the threshold, this RoI will be regarded as background and discarded, consistent with the process in~\cite{yan2019meta,Xiao2020FSDetView}.

\subsection{Training Procedure}\label{M_C}

{\bfseries Training Phase.}
Following the common practice in~\cite{kang2019few,yan2019meta,wang2020frustratingly,Xiao2020FSDetView}, we train our network in two phases. Base data $D_{\text {\textit{base}}}$ from base classes $C_{\text {\textit{base}}}$ with abundant samples per class are used to train the model in the first base training phase. Then the balanced base data $D_{\text {\textit{base}}}$ and novel data $D_{\text {\textit{novel}}}$  with only $K$ samples per class are fed into the network in the second fine-tuning phase for the fast adaptation to novel classes.

{\bfseries Training Data Organization.}
In the training phase, distinct from general object detectors that applying an image $x_i$ as the training mini-batch, our few-shot detector applies a task $T_{i}$ as an input training data in the meta-learning paradigm~\cite{kang2019few,yan2019meta,Xiao2020FSDetView}. Each input task $T_i=S_i \cup Q_i$ is the union of a query set $Q_i$ and a support set $S_i$, where $Q_i$ provided to query branch in Fig.\ref{fig:2} is a query image $q_i$ containing objects from $m$ classes $C_{i}^{\text {\textit{meta}}} \subseteq C_{\text {\textit{base}}} \cup C_{\text {\textit{novel}}}$, and $S_i$ for support branch contains $m$ $K$-shot ($K=200$ in the base training phase and $K$ = $\{1, 2, 3, 5, 10\}$ in the fine-tuning phase) clusters $\{g_{i}^{p}\}_{p=1}^{m}$, where each cluster $g_{i}^{p}$ includes $K$ images $\{s_{i, j}^{p}\}_{j=1}^{K}$ in the category $p$. Each support image $s_{i, j}^{p}$ is depth-wise concatenated with a structured binary $\textit{mask}_{i, j}^{p}$ (see the input support set in Fig.~\ref{fig:2}, only one object is considered when multiple objects are present within an image).

{\bfseries Loss Function.}
We optimize our network in both training phases using the loss function introduced in~\cite{yan2019meta}:
\begin{equation} \label{eq:6}
	L=L_{\textit{Faster R-CNN}}+L_{\textit{meta}}
\end{equation}
where $L_{\textit{Faster R-CNN}}$ denotes the loss function of base detector Faster R-CNN, involving the RPN loss, and classification loss together with bounding box regression loss for the state estimation, $L_{\textit{meta}}$ denotes a meta loss aiming at encouraging class-attentive vectors of support set to distinguish with each other.

Since our proposed Dual Attention Generator and Dual Query Encoder are in a dual architecture, $L_{\textit{meta}}$ in this paper is a combination of two components for classification and regression, respectively. Thus the loss function in Eq.~(\ref{eq:6}) can be precisely written as:
\begin{equation}
	L=L_{\textit{Faster R-CNN}}+L_{\textit{meta-cls}}+L_{\textit{meta-reg}}
\end{equation}
where $L_{\textit{meta-cls}}$ and $L_{\textit{meta-reg}}$ denote the meta losses for classification and boundinng box regression, respectively, implemented by the cross-entropy loss.

\begin{table*}
	\small
	%\vspace{-0.2cm}
	\caption{\small Few-shot detection performance ($AP_{50}$) for novel categories on PASCAL VOC dataset. We evaluate baselines on three different novel sets. Our approach consistently outperforms other methods by a large margin. $^{*n}$Reported results are averaged over $n$ repeated runs.}
	\vspace{-0.2cm}
	\label{tab:1}
	\centering
	{\begin{tabular}{c|l|l|l|ccccc|ccccc|ccccc}
			\hline
			%\toprule[1.1pt]
			\multicolumn{4}{c|}{}                                & \multicolumn{5}{c|}{Novel Set 1}                                              & \multicolumn{5}{c|}{Novel Set 2}                                              & \multicolumn{5}{c}{Novel Set 3}                                              \\
			\multicolumn{4}{c|}{\multirow{-2}{*}{Method / Shot}} & 1             & 2             & 3             & 5             & 10            & 1             & 2             & 3             & 5             & 10            & 1             & 2             & 3             & 5             & 10            \\ %\midrule[0.9pt]
			\hline
			\multicolumn{4}{c|}{LSTD \cite{chen2018lstd}}                   & 8.2           & 11.0           & 12.4          & 29.1          & 38.5          & 11.4          & 3.8           & 5.0           & 15.8          & 31.0          & 12.6          & 8.5           & 15.0          & 27.3          & 36.3          \\
			\multicolumn{4}{c|}{FSRW \cite{kang2019few}}                   & 14.8          & 15.5          & 26.7          & 33.9          & 47.2          & 15.7          & 15.2          & 22.7          & 30.1          & 40.5          & {21.3}          & 25.6          & 28.4          & 42.8          & 45.9          \\
			\multicolumn{4}{c|}{MetaDet$^{*5}$ \cite{wang2019meta}}                & 18.9          & 20.6          & 30.2          & 36.8          & 49.6          & {21.8}          & 23.1          & 27.8          & 31.7          & 43.0          & 20.6          & 23.9          & 29.4          & 43.9          & 44.1          \\
			\multicolumn{4}{c|}{Meta R-CNN$^{*5}$ \cite{yan2019meta}}             & 19.9          & 25.5          & 35.0          & 45.7          & 51.5          & 10.4          & 19.4          & 29.6          & 34.8          & 45.4          & 14.3          & 18.2          & 27.5          & 41.2          & 48.1          \\
			\multicolumn{4}{c|}{TFA$^{*30}$ w/ fc \cite{wang2020frustratingly}}               & 22.9          & 34.5          & 40.4          & 46.7          & 52.0          & 16.9          & 26.4          & 30.5          & 34.6          & 39.7          & 15.7          & 27.2          & 34.7          & 40.8          & 44.6          \\
			\multicolumn{4}{c|}{TFA$^{*30}$ w/ cos \cite{wang2020frustratingly}}              & {25.3}          & {36.4}          & 42.1          & 47.9          & 52.8          & 18.3          & {27.5}          & 30.9          & 34.1          & 39.5          & 17.9          & 27.2          & 34.3          & 40.8          & 45.6          \\
			\multicolumn{4}{c|}{FSDetView$^{*10}$ \cite{Xiao2020FSDetView}}              & 24.2          & 35.3          & {42.2}          & {49.1}          & {57.4}          & {21.6}          & 24.6          & {31.9}          & {37.0}          & {45.7}          & 21.2          & {30.0}          & {37.2}          & {43.8}          & {49.6}          \\ %\midrule[0.9pt]
			
			\multicolumn{4}{c|}{AFD-Net$^{*30}$ (Ours)}    & \textbf{31.7} & \textbf{41.4} & \textbf{49.5} & \textbf{54.6} & \textbf{60.3} & \textbf{23.2} & \textbf{31.3} & \textbf{38.4} & \textbf{41.9} & \textbf{46.9} & \textbf{27.4} & \textbf{35.3} & \textbf{41.7} & \textbf{46.7} & \textbf{53.5} \\ \hline%\bottomrule[1.1pt]
	\end{tabular}}
	%\vspace{-0.2cm}
\end{table*}

\begin{table*}\centering
	\small
	%\vspace{0.5cm}
	\caption{\small Few-shot detection performance ($AP_{50}$) for base and novel categories on Novel Set 1 of PASCAL VOC dataset. Our approach outperforms other methods on both base and novel classes. $^{*n}$Reported results are averaged over $n$ repeated runs.}
	%\vspace{0.15cm}
	\vspace{-0.2cm}
	\label{tab:2}
	\setlength{\tabcolsep}{0.8pt}
	%\resizebox{500pt}{67pt}
	%\fontsize{7.6}{10.5}
	{\begin{tabular}{c|c|cccccc|cccccccccccccccc|c}
			\hline
			\multirow{2}{*}{Shot} & \multirow{2}{*}{Method} & \multicolumn{6}{c|}{Novel classes} & \multicolumn{16}{c|}{Base classes}                       & \multirow{2}{*}{mAP} \\
			&                        & bird & bus & cow & mbike & sofa & mean & aero   & bike & boat & bottle & car & cat & chair & table & dog & horse & person & plant & sheep & train & tv 
			& mean  &                      \\ \hline
			\multirow{4}{*}{3}     & LSTD \cite{chen2018lstd}                      & 23.1   & 22.6   & 15.9   & 0.4   & 0.0   & 12.4  & \textbf{74.8} & 68.7 & 57.1 & 44.1 & 78.0 & 83.4 & 46.9 & 64.0 & 78.7 & \textbf{79.1} & 70.1 & 39.2 & 58.1 & \textbf{79.8} & \textbf{71.9} & 66.3 & 52.8                  \\
			& FSRW \cite{kang2019few}                     &26.1 &19.1 &40.7 &20.4 &{27.1} &26.7 &{73.6} &{73.1} &56.7 &41.6 &{76.1} &78.7 &42.6 &{\textbf{66.8}} &72.0 &77.7 &68.5 &{\textbf{42.0}} &57.1 &{74.7} &70.7 &{64.8} &55.2                  \\
			& Meta R-CNN$^{*5}$ \cite{yan2019meta}                      &{30.1} &{44.6} &\textbf{50.8} &{38.8} &10.7 &{35.0} &67.6 &70.5 &{\textbf{59.8}} &{50.0} &75.7 &81.4 &{44.9} &57.7 &{76.3} &74.9 &{76.9} &34.7 &58.7 &{74.7} &67.8 &{64.8} &{57.3}                  \\
			& AFD-Net$^{*30}$ (Ours)                      & \textbf{51.8}   & \textbf{60.3}   & 43.8   & \textbf{60.5}   & \textbf{31.3}   & \textbf{49.5}  & 69.9 & \textbf{75.2} & 56.9 & \textbf{57.9} & \textbf{79.5} & \textbf{84.2} & \textbf{47.9} & 60.5 & \textbf{82.4} & 76.7 & \textbf{77.4} & 40.8 & \textbf{68.6} & 75.2 & 71.0 & \textbf{68.3} & \textbf{63.6}                  \\
			\hline
			\multirow{4}{*}{10}    & LSTD \cite{chen2018lstd}                      & 22.8   & 52.5   & 31.3   & 45.6   & 40.3   & 38.5  & 70.9 & 71.3 & 59.8 & 41.1 & 77.1 & 81.9 & 45.1 & \textbf{67.2} & 78.0 & 78.9 & 70.7 & 41.6 & 63.8 & \textbf{79.7} & 66.8 & 66.3 & 59.4                  \\
			& FSRW \cite{kang2019few}                      &30.0 &{62.7} &43.2 &{60.6} &39.6 &{47.2} &65.3 &73.5 &54.7 &39.5 &75.7 &81.1 &35.3 &62.5 &72.8 &78.8 &68.6 &{41.5} &59.2 &{76.2} &69.2 &63.6 &{59.5}                  \\
			& Meta R-CNN$^{*5}$ \cite{yan2019meta}                      &{52.5} &{55.9} &{52.7} &{54.6} &{41.6} &{51.5} &68.1 &73.9 &{59.8} &54.2 &{80.1} &{82.9} &{48.8} &{62.8} &{80.1} &{\textbf{81.4}} &{77.2} &37.2 &65.7 &75.8 &{70.6} &{67.9} &{63.8}                  \\
			& AFD-Net$^{*30}$ (Ours)                      & \textbf{62.7}   & \textbf{67.9}   & \textbf{60.2}   & \textbf{68.1}   & \textbf{42.7}   & \textbf{60.3}  & \textbf{73.0} & \textbf{77.9} & \textbf{60.4} & \textbf{59.7} & \textbf{81.8} & \textbf{85.4} & \textbf{51.0} & 66.5 & \textbf{84.7} & 80.3 & \textbf{78.1} & \textbf{44.6} & \textbf{70.1} & 77.6 & \textbf{72.7} & \textbf{70.9} & \textbf{68.3}                  \\
			\hline
	\end{tabular}}
	%\vspace{-0.2cm}
\end{table*}

\section{Experiments}
\label{sec:Experiments}
\subsection{Experimental Settings}\label{E_A}

{\bfseries Benchmarks.}
We evaluate our method on general object detection benchmarks, \emph{i.e.}, PASCAL VOC 2007, 2012 and MS COCO, in few-shot detection settings. As for PASCAL VOC, we follow the common practice in previous work~\cite{kang2019few,yan2019meta,wang2020frustratingly,Xiao2020FSDetView}, and use train/val sets of VOC 07 and 12 for training while the test set of VOC 07 for testing. This benchmark covers 20 object categories, where 15 of them are regarded as base classes for the base training phase and the remaining 5 categories with only $K$ ($K=\{1, 2, 3, 5, 10\}$) samples per class are considered as novel classes for few-shot fine-tuning phase. For the fair quantitative comparison, we adopt the same three class splits provided in~\cite{kang2019few}. For the evaluation protocol, we use mean Average Precision ($mAP$) of novel objects and the Intersection of Union (\textit{IoU}) is set as 0.5 ($AP_{50}$). Another benchmark we evaluate on is MS COCO, which has 80k train images and 40k validation images, covering 80 classes. Among them, we denote the 20 classes overlapped with PASCAL VOC as novel classes with $K$ ($K=\{10, 30\}$) samples per category and the rest 60 classes as base classes. We use 5k images from validation set for evaluation with the standard COCO-style evaluation metrics~\cite{liu2016ssd,redmon2018yolov3} and the rest for training. To compare empirically, on both datasets, we evaluate on novel classes over $n$ ($n=30$ for PASCAL VOC and $n=10$ for MS COCO) repeated runs unless otherwise specified, and report the average performance.

{\bfseries Implementation Details.}
Our proposed approach is implemented in the \textit{PyTorch}~\cite{paszke2017automatic} library. It employs the backbone network of ResNet~\cite{he2016deep} pre-trained on ImageNet~\cite{russakovsky2015imagenet} along with a RoIAlign~\cite{he2017mask} layer. Specifically, we use ResNet-101 on PASCAL VOC and ResNet-50 on MS COCO. Our model is trained end-to-end using a batch size of 4 on a single Nvidia GeForce GTX 1080Ti with 13 GB memory. For optimization, we keep the setups in~\cite{yan2019meta,Xiao2020FSDetView} on both benchmarks. Concretely, we adopt the SGD optimizer with an initial learning rate of 0.001, weight decay of 0.0005 and momentum of 0.9. The model is trained for 20 epochs in the base training phase and 9 epochs in the $K$-shot fine-tuning phase. The learning rate is reduced by 0.1 every 5 epochs in two training phases. During inference, the support branch will be removed for directly performing detection without requiring support images as input. It is because that class-attentive vectors for  model reweighting can be obtained in the $K$-shot fine-tuning phase by averaging the outputs from support branch over the input $K$ support images.

\vspace{-0.2cm}
\subsection{Comparison with State-of-the-art Methods}\label{E_B}
{\bfseries PASCAL VOC.}
Our evaluation results are presented in Table \ref{tab:1}. The comparison experiments cover few-shot object detection scenarios in five setups ($K=\{1, 2, 3, 5, 10\}$) across three base/novel set splits. As can be observed, our method outperforms recent state-of-the-art methods by a large margin and obtains the best performance in all 15 cases. We notice that in the majority of cases, the improvements are much larger than the gap among previous approaches, which indicates the strong generalization ability of our model. Surprisingly, although the existence of high variance of support data in the extreme few-shot setup ($K=1$), we obtain large improvements (+6.4\% in the first split and +6.2\% in the third split), showing the robustness of our model in tough scenarios. Besides, in the robust 10-shot setup, the improvements (+2.9\% in the first split, +1.2\% in the second split and +3.9\% in the third split) are lower than those in other setups. This can be explained by that, as the number of instances per novel class increases, the sample variance decreases and the detection performance stabilizes.

Taking evaluation performance on base classes into consideration, we provide detailed results on the first base/novel split in Table \ref{tab:2}. Note that our proposed method achieves a new state-of-the-art performance on both base and novel classes. Additionally, the performance improvements are much larger in novel classes compared with that in base classes. As for the performance on the single category, our approach has the best detection performance for most of categories except several base classes in 3-shot scenario, such as ``table'' and ``train''. We will give detailed analysis in the ablation study, from the perspective of sample variance.

\begin{table*}[] \centering 
	\small
	%\resizebox{445pt}{125pt}
	%\vspace{-0.2cm}
	\caption{\small Few-shot detection performance for novel categories on MS COCO dataset. Our approach achieves an significant improvement over other methods with notably smaller standard deviations. $^{*n}$Reported results are averaged over $n$ repeated runs.}
	\vspace{-0.2cm}
	\setlength{\tabcolsep}{0.8mm}
	{\begin{tabular}{c|c|cccccc|cccccc}
			\hline
			%\toprule[1.1pt]
			&                              & \multicolumn{6}{c|}{Average Precision}                                                                                                                                & \multicolumn{6}{c}{Average Recall}                                                                                                                                   \\
			\multirow{-2}{*}{Shots} & \multirow{-2}{*}{Method}     & 0.5:0.95                  & 0.5                       & 0.75                      & S                         & M                         & L                         & 1                         & 10                        & 100                       & S                         & M                         & L                         \\ \hline%\midrule[0.9pt]
			& LSTD \cite{chen2018lstd}                & 3.2                       & 8.1                       & 2.1                       & 0.9                       & 2.0                       & 6.5                       & 7.8                       & 10.4                      & 10.4                      & 1.1                       & 5.6                       & 19.6                      \\
			& FSRW \cite{kang2019few}                & 5.6                       & 12.3                      & 4.6                       & 0.9                       & 3.5                       & 10.5                      & 10.1                      & 14.3                      & 14.4                      & 1.5                       & 8.4                       & 28.2                      \\
			& MetaDet$^{*5}$ \cite{wang2019meta}             & 7.1                       & 14.6                      & 6.1                       & 1.0                       & 4.1                       & 12.2                      & 11.9                      & 15.1                      & 15.5                      & 1.7                       & 9.7                       & 30.1                      \\
			& Meta R-CNN$^{*5}$ \cite{yan2019meta}          & 8.7                       & 19.1                      & 6.6                       & 2.3                       & 7.7                       & 14.0                      & 12.6                      & 17.8                      & 17.9                      & 7.8                       & 15.6                      & 27.2                      \\
			& TFA$^{*10}$ w/ fc \cite{wang2020frustratingly}            & 9.1 $\pm$ 0.5                       & 17.3$\pm$1.0                      & 8.5$\pm$0.5                       & -                         & -                         & -                         & -                         & -                         & -                         & -                         & -                         & -                         \\
			& TFA$^{*10}$ w/ cos \cite{wang2020frustratingly}           & 9.1 $\pm$ 0.5                      & 17.1$\pm$1.1                      & 8.8$\pm$0.5                       & -                         & -                         & -                         & -                         & -                         & -                         & -                         & -                         & -                         \\
			& FSOD \cite{fan2020few}           & 11.1                      & 20.4                      & 10.6                       & -                       & -                      & -                      & -                      & -                      & -                      & -                       & -                      & -                      \\
			& FSDetView$^{*10}$ \cite{Xiao2020FSDetView}           & 12.5                      & 27.3                      & 9.8                       & 2.5                       & 13.8                      & 19.9                      & 20.0                      & 25.5                      & 25.7                      & 7.5                       & 27.6                      & 38.9                      \\
			\multirow{-8}{*}{10}    & AFD-Net$^{*10}$ (Ours) & \textbf{17.3 $\pm$ 0.1} & \textbf{33.4 $\pm$ 0.2} & \textbf{16.6 $\pm$0.1} & \textbf{5.7$\pm$0.2} & \textbf{18.6$\pm$0.2} & \textbf{27.4$\pm$0.2} & \textbf{24.5$\pm$0.1} & \textbf{31.5$\pm$0.1} & \textbf{31.8$\pm$0.1} & \textbf{12.7$\pm$0.2} & \textbf{33.8$\pm$0.1} & \textbf{45.8$\pm$0.2} \\ \hline %\midrule[0.7pt]
			& LSTD \cite{chen2018lstd}                & 6.7                       & 15.8                      & 5.1                       & 0.4                       & 2.9                       & 12.3                      & 10.9                      & 14.3                      & 14.3                      & 0.9                       & 7.1                       & 27.0                      \\
			& FSRW \cite{kang2019few}                & 9.1                       & 19.0                      & 7.6                       & 0.8                       & 4.9                       & 16.8                      & 13.2                      & 17.7                      & 17.8                      & 1.5                       & 10.4                      & 33.5                      \\
			& MetaDet$^{*5}$ \cite{wang2019meta}             & 11.3                      & 21.7                      & 8.1                       & 1.1                       & 6.2                       & 17.3                      & 14.5                      & 18.9                      & 19.2                      & 1.8                       & 11.1                      & 34.4                      \\
			& Meta R-CNN$^{*5}$ \cite{yan2019meta}          & 12.4                      & 25.3                      & 10.8                      & 2.8                       & 11.6                      & 19.0                      & 15.0                      & 21.4                      & 21.7                      & 8.6                       & 20.0                      & 32.1                      \\
			& TFA$^{*10}$ w/fc \cite{wang2020frustratingly}            & 12.0$\pm$0.4                      & 22.2$\pm$0.6                      & 11.8$\pm$0.4                      & -                         & -                         & -                         & -                         & -                         & -                         & -                         & -                         & -                         \\
			& TFA$^{*10}$ w/cos \cite{wang2020frustratingly}           & 12.1$\pm$0.4                      & 22.0$\pm$0.7                      & 12.0$\pm$0.5                      & -                         & -                         & -                         & -                         & -                         & -                         & -                         & -                         & -                         \\
			& FSDetView$^{*10}$ \cite{Xiao2020FSDetView}           & 14.7                      & 30.6                      & 12.2                      & 3.2                       & 15.2                      & 23.8                      & 22.0                      & 28.2                      & 28.4                      & 8.3                       & 30.3                      & 42.1                      \\
			\multirow{-8}{*}{30}    & AFD-Net$^{*10}$ (Ours) & \textbf{19.1$\pm$0.04} & \textbf{35.8$\pm$0.1} & \textbf{18.7$\pm$0.2} & \textbf{5.7$\pm$0.1} & \textbf{20.6$\pm$0.1} & \textbf{29.2$\pm$0.2} & \textbf{26.0$\pm$0.2} & \textbf{33.4$\pm$0.2} & \textbf{33.7$\pm$0.2} & \textbf{14.3$\pm$0.1} & \textbf{36.1$\pm$0.2} & \textbf{47.6$\pm$0.4} \\ \hline %\bottomrule[1.1pt]
	\end{tabular}}
	%\setlength{\belowcaptionskip}{-0.9cm}
	%\vspace{0.3cm}
	%\vspace{-0.1cm}
	\label{tab:3}
	%\vspace{0.05cm}
\end{table*}

{\bfseries MS COCO.}
We show the evaluation of 20 novel classes on MS COCO in 10/30-shot setups, and report the standard COCO-style metrics in Table~\ref{tab:3}. Obviously, our approach significantly outperforms recent state-of-the-art methods with much smaller confidence intervals, despite the complexity and a huge amount of data in MS COCO, verifying its effectiveness and robustness. Note that our method performs much better in detection of small objects in comparison with other state-of-the-arts, indicating that our proposed network indeed enhances feature representations and obtains high-quality image information.

In Fig.~\ref{fig:5}, we provide the comparison of qualitative 30-shot detection results between our approach and baseline method FSDetView~\cite{Xiao2020FSDetView}. We can find that our method performs better on detecting small and occluded objects, \emph{e.g.}, the first two columns. This is consistent with the observations from Table~\ref{tab:3}. Besides, our model provides accurate locations of target objects, \emph{e.g.}, the third column. The last two columns show the ability of our approach in classifying objects from both base and novel categories, including correctly classifying target objects and avoiding duplicate estimations of object categories.

%-------------------------------------------------------------------------

{% \renewcommand{\arraystretch}{0}
	\begin{figure*}[!ht]
		\centering
		\footnotesize
		%\flushleft
		\vspace{-0.2cm}
		\setlength{\tabcolsep}{0.1em}
		\adjustbox{width=.9\linewidth}{
			\begin{tabular}{ccccccc}
				\multirow{2}{*}{\rotatebox{90}{\hspace{4mm}Base Classes}}& 
				\rotatebox{90}{\hspace{4mm}Baseline} &
				\includegraphics[height=0.8in]{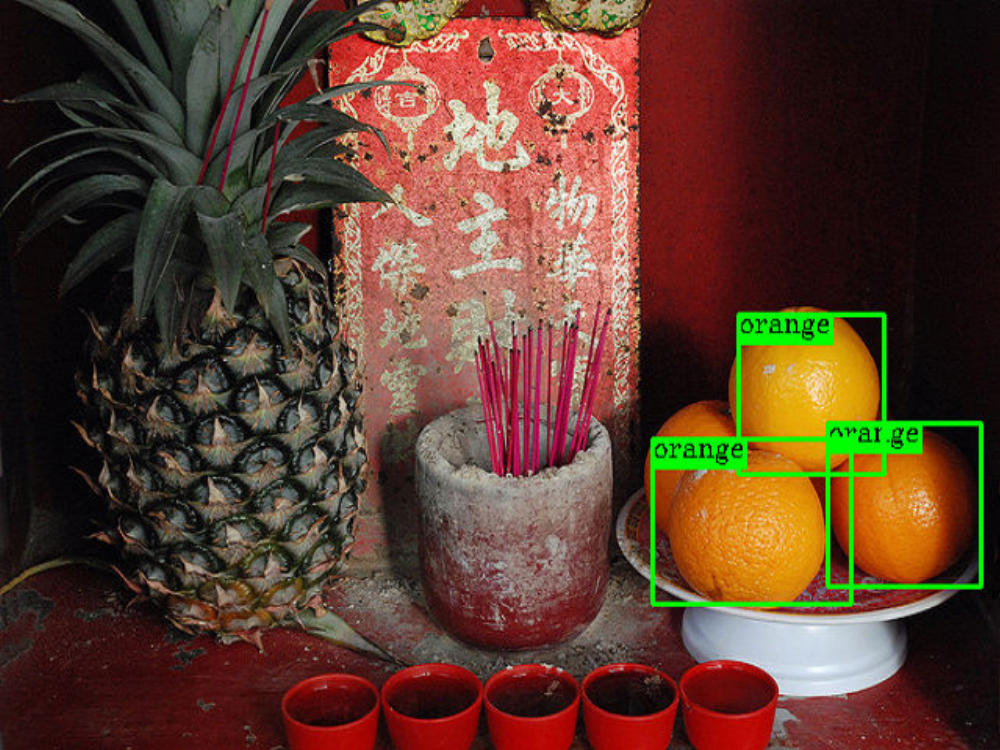} &  \includegraphics[height=0.8in]{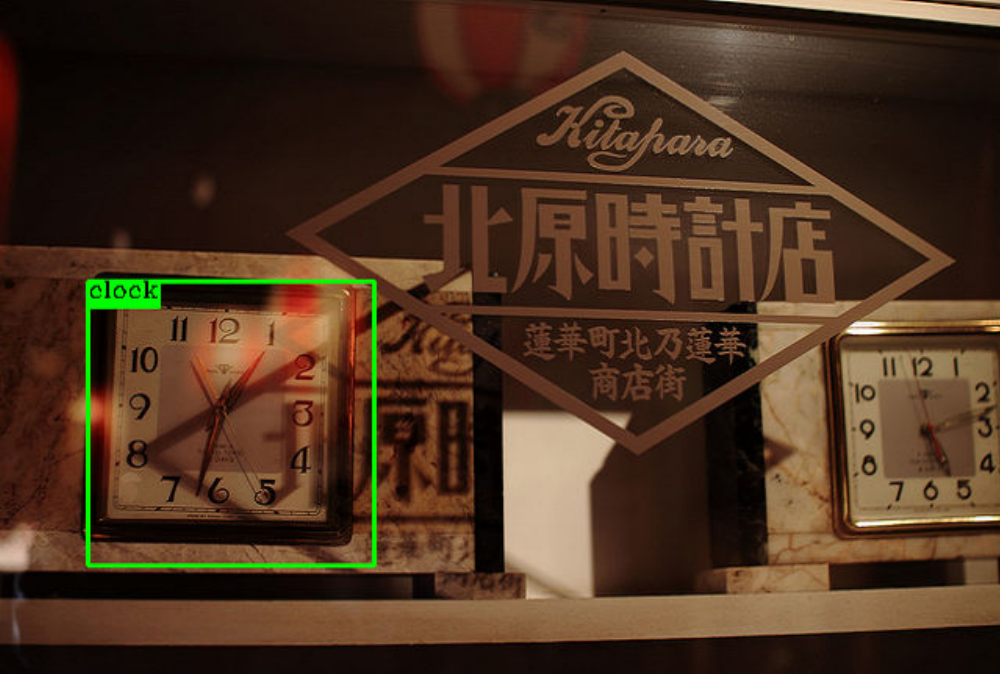} &
				\includegraphics[height=0.8in]{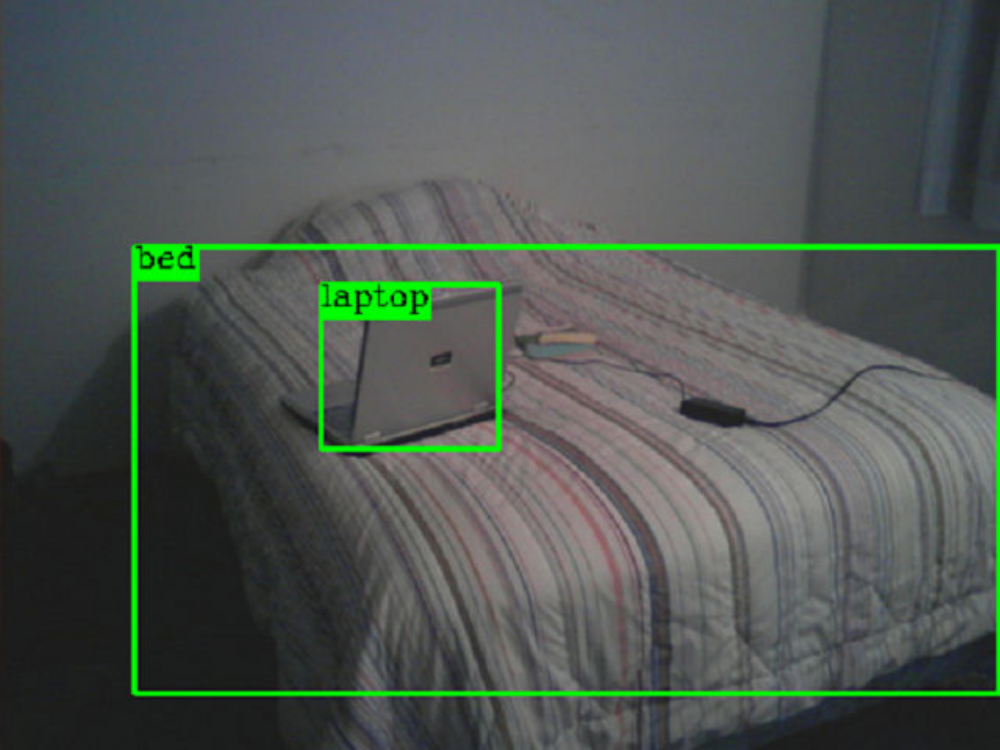} &
				\includegraphics[height=0.8in]{figures/base/baseline/_wrong_205776_det.pdf} & \includegraphics[height=0.8in]{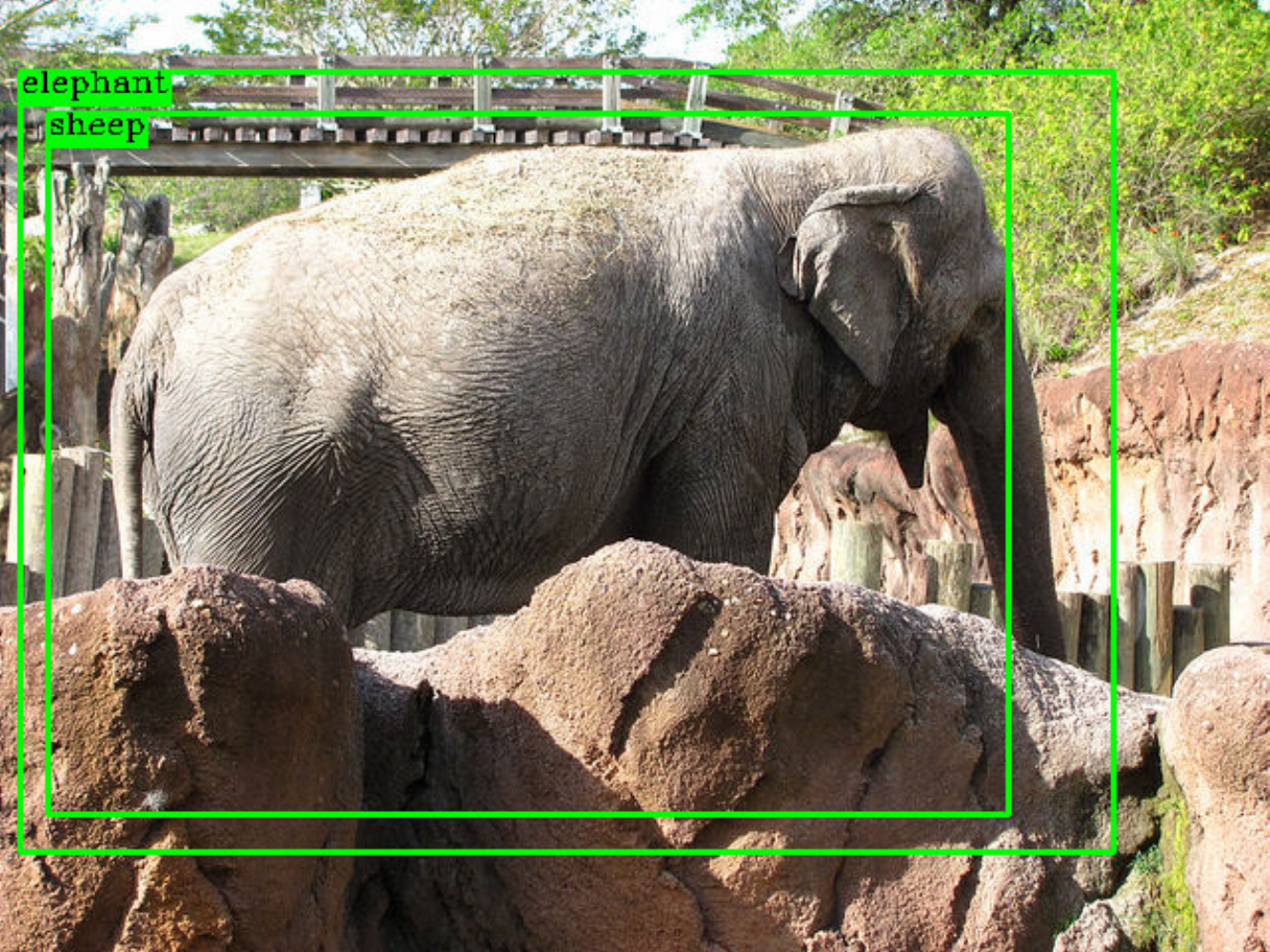} \\
				& \rotatebox{90}{\hspace{4mm}Ours} &
				\includegraphics[height=0.8in]{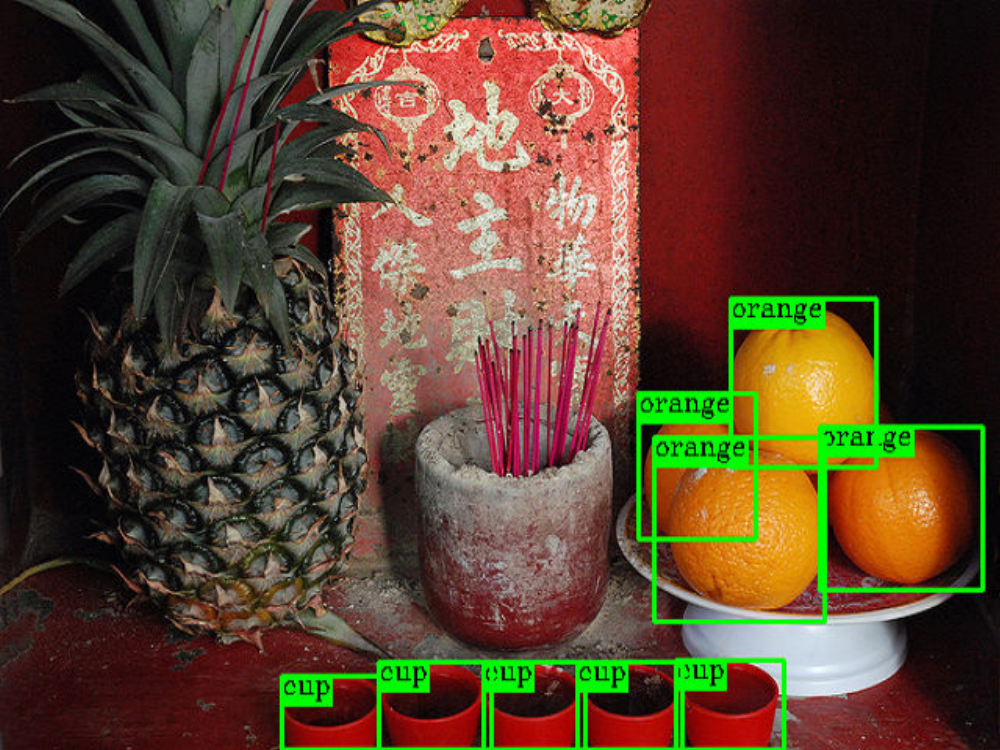} &  \includegraphics[height=0.8in]{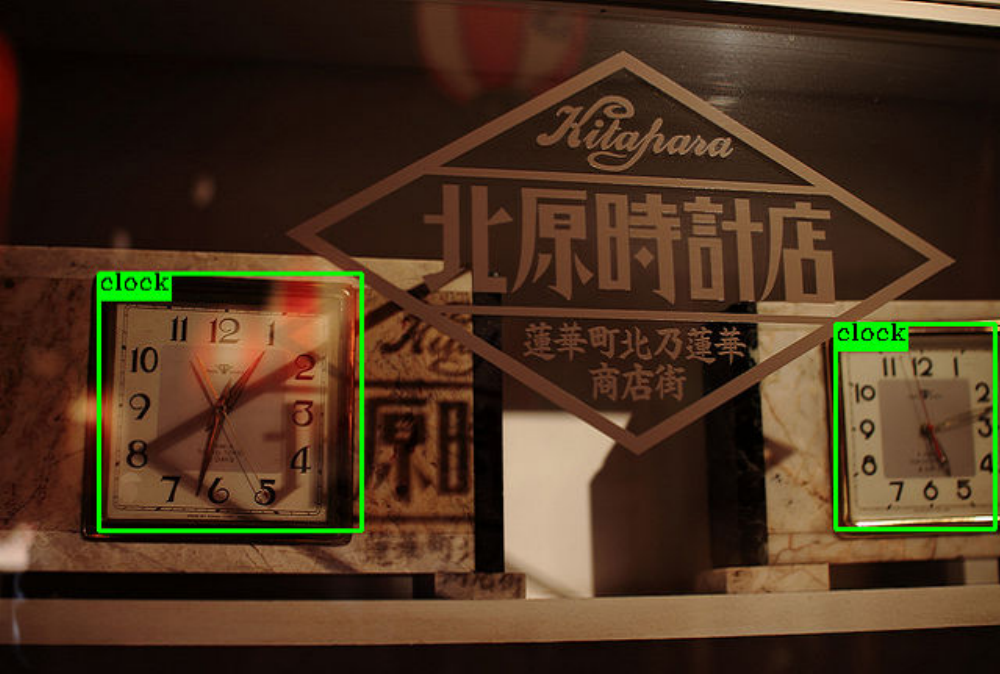} &
				\includegraphics[height=0.8in]{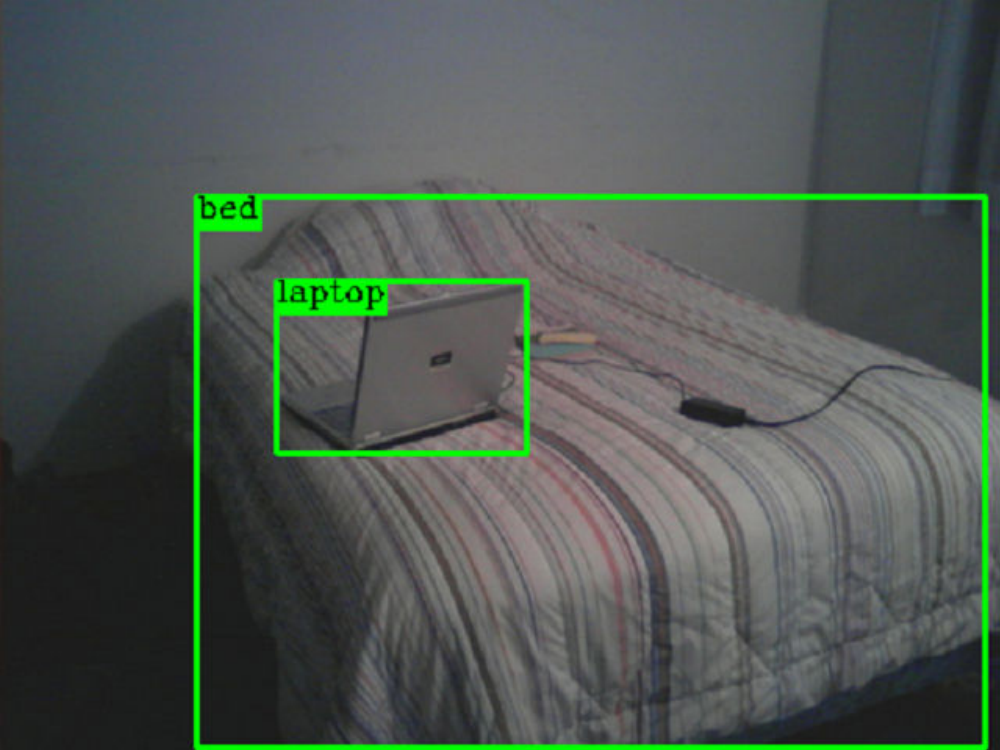} &
				\includegraphics[height=0.8in]{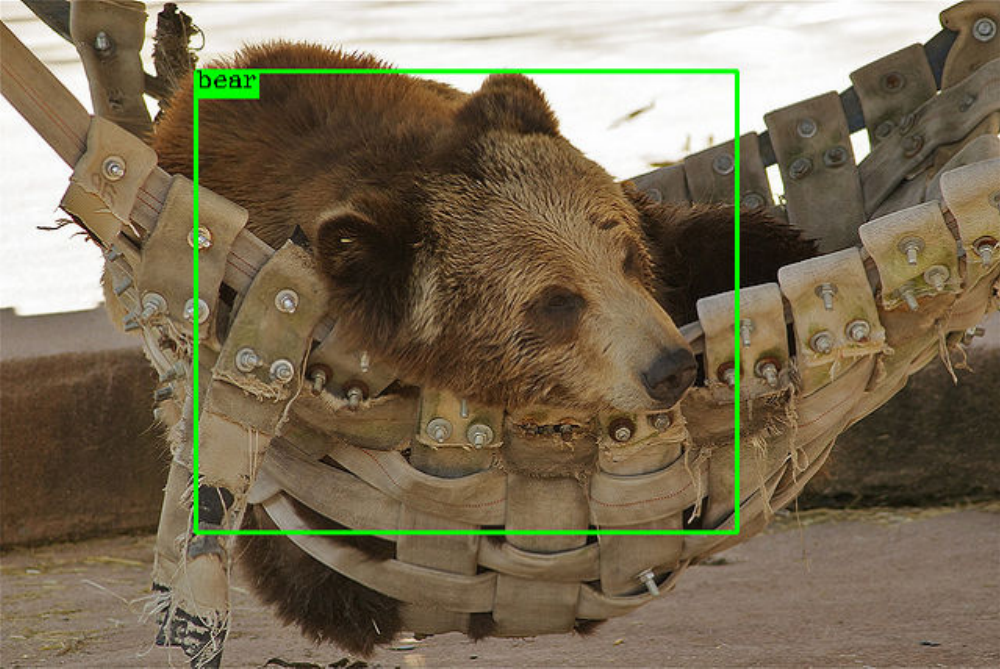} & \includegraphics[height=0.8in]{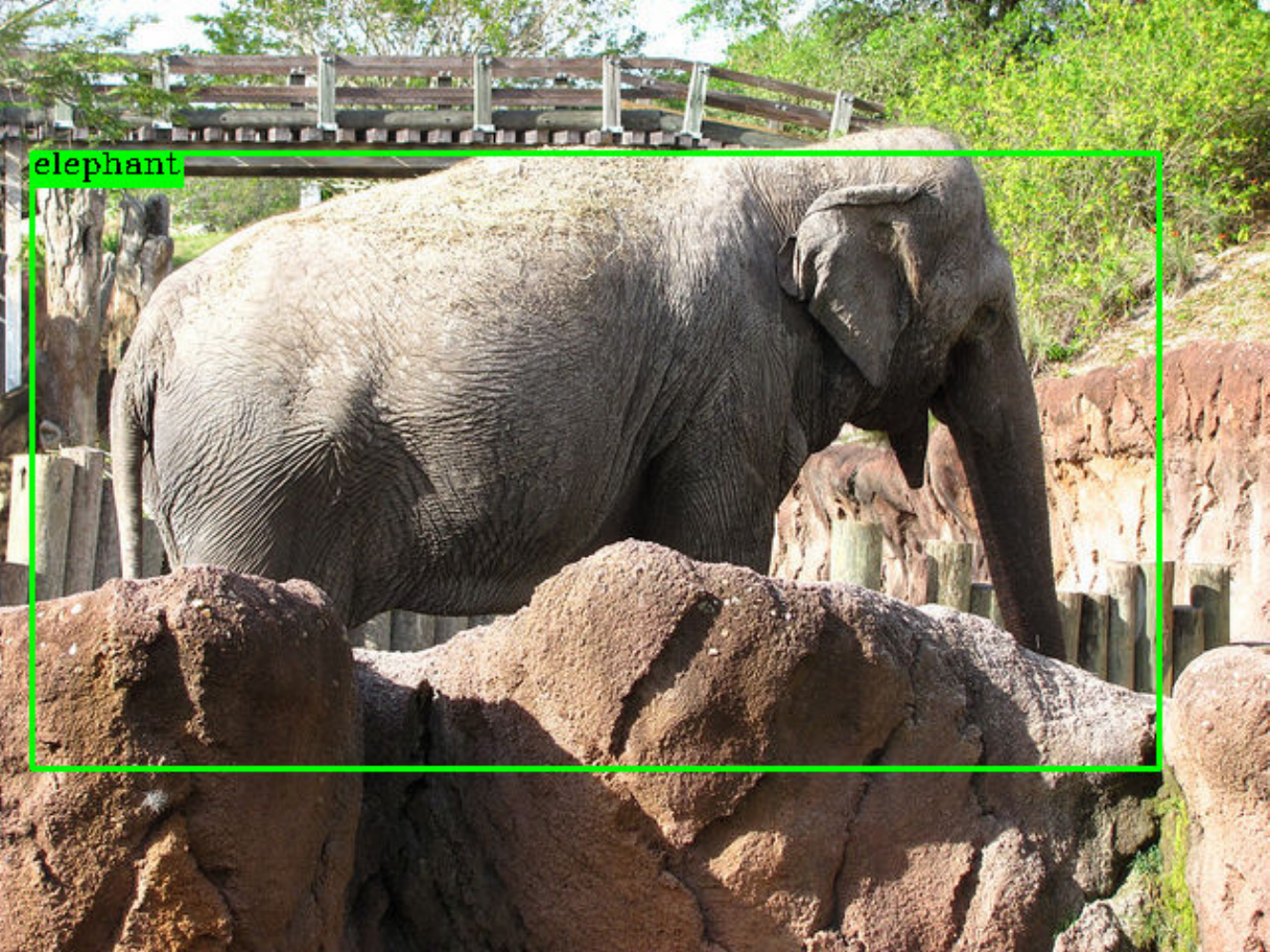} \\
				\multirow{2}{*}{\rotatebox{90}{Novel Classes}} &
				\rotatebox{90}{\hspace{4mm}Baseline} &
				\includegraphics[height=0.8in]{figures/novel/baseline/_missed_195842_det.pdf} & \includegraphics[height=0.8in]{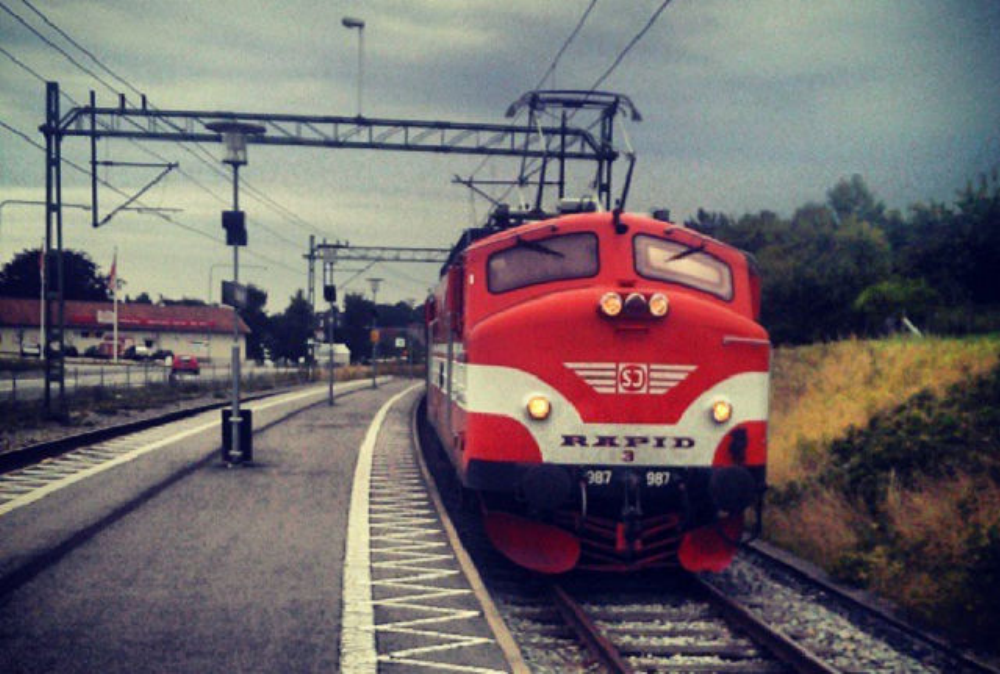} &
				\includegraphics[height=0.8in]{figures/novel/baseline/_mislocalizing_516143_det.pdf} & \includegraphics[height=0.8in]{figures/novel/baseline/_wrong_545958_det.pdf} & \includegraphics[height=0.8in]{figures/novel/baseline/_duplicate_30494_det.pdf} \\
				& \rotatebox{90}{\hspace{4mm}Ours} &
				\includegraphics[height=0.8in]{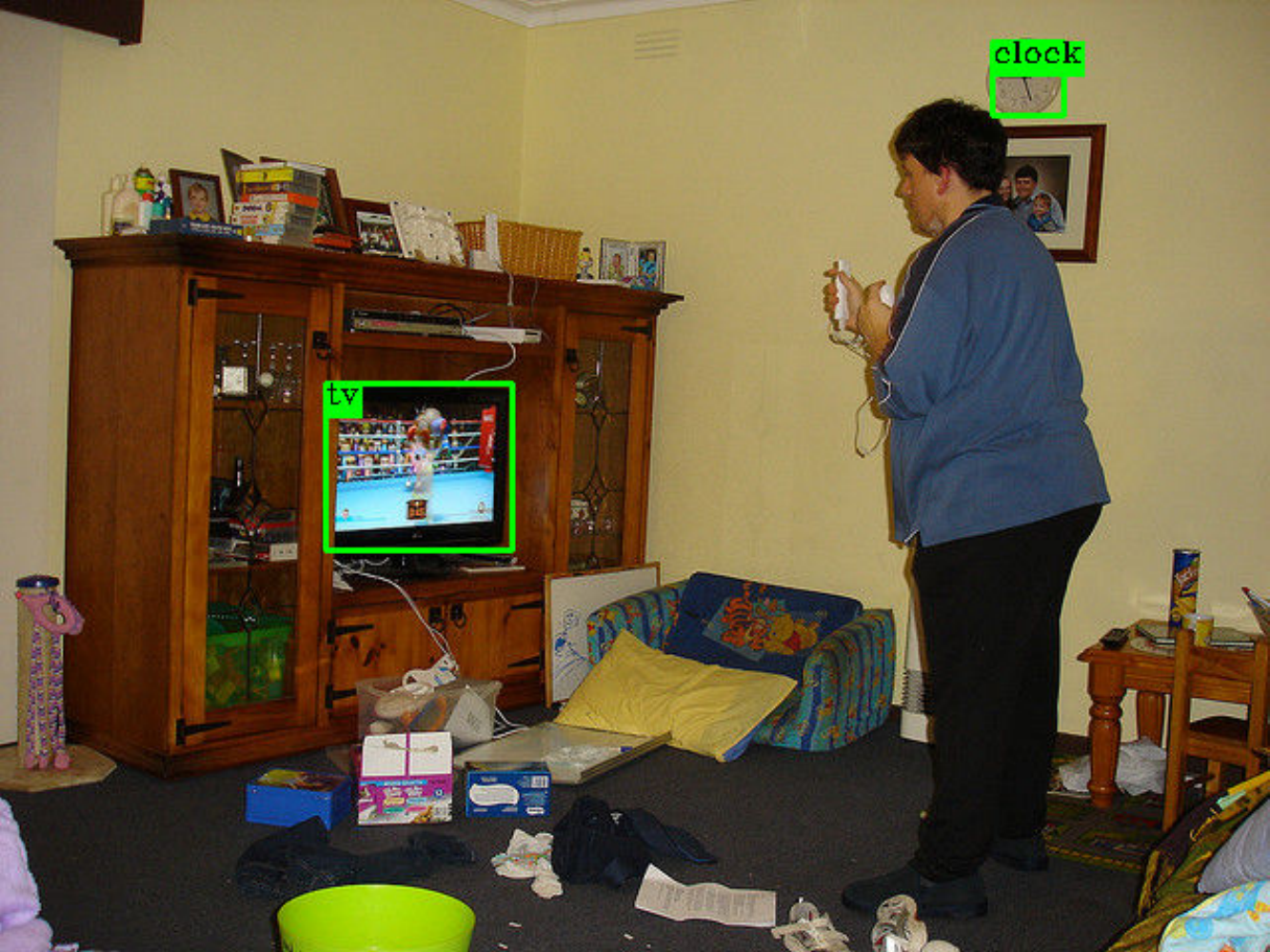} & \includegraphics[height=0.8in]{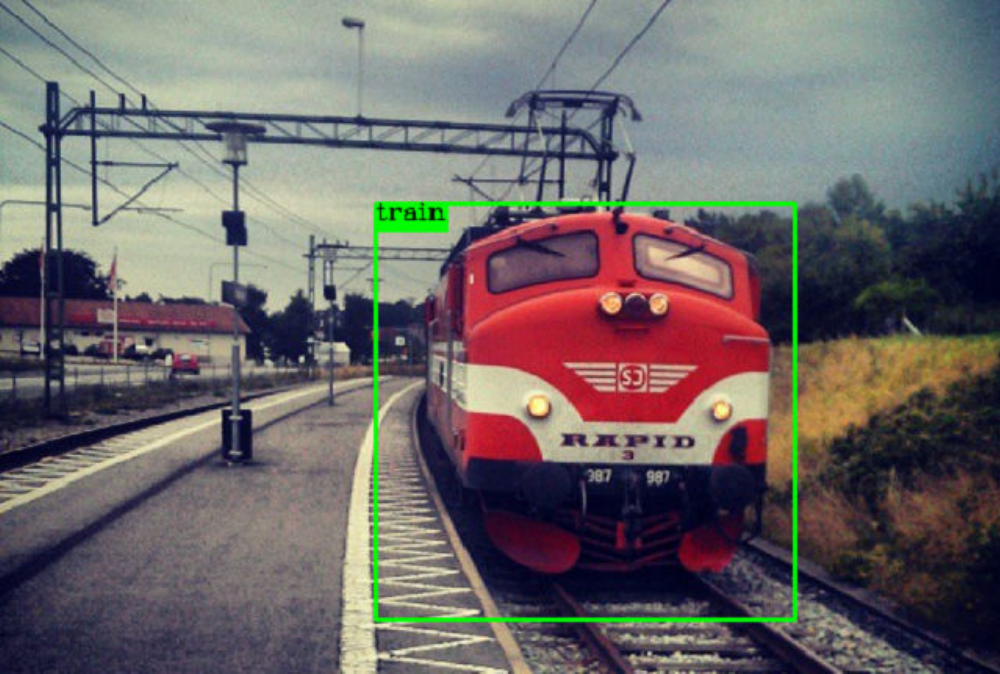} &
				\includegraphics[height=0.8in]{figures/novel/ours/_mislocalizing_516143_det.pdf} & \includegraphics[height=0.8in]{figures/novel/ours/_wrong_545958_det.pdf} & \includegraphics[height=0.8in]{figures/novel/ours/_duplicate_30494_det.pdf} \\
		\end{tabular}}
		\vspace{-0.2cm}
		\caption{\small Comparison of qualitative 30-shot detection results on base and novel classes from MS COCO between our approach and baseline method FSDetView~\cite{Xiao2020FSDetView}.
		}
		\vspace{-0.2cm}
		\label{fig:5}
\end{figure*}}

%--------------------------------------------------------------------------
\begin{figure}[t] \centering
	\vspace{-0.1cm}
	\begin{center}
		\includegraphics[width=1.0\linewidth]{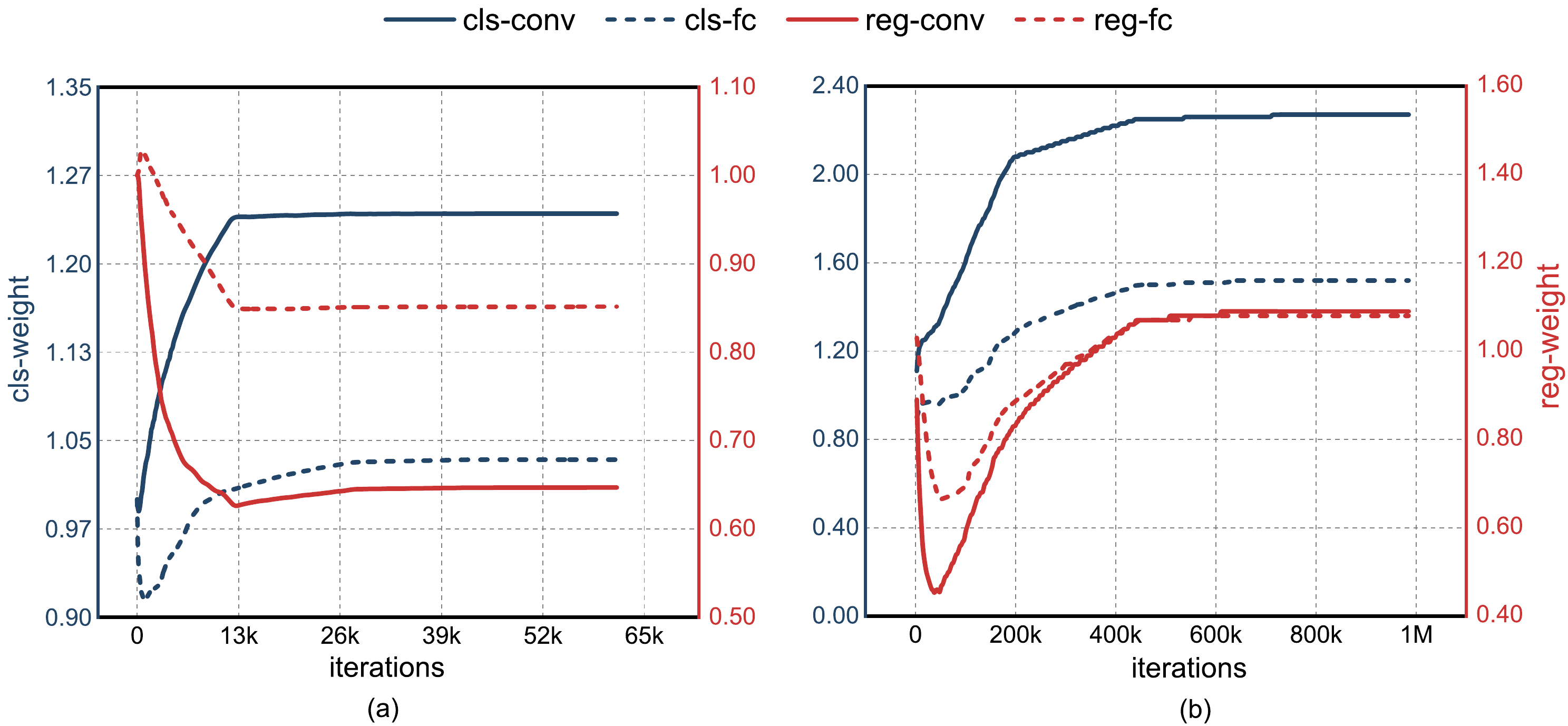}
	\end{center}
	\vspace{-0.25cm}
	\caption{\small Performance visualization of our proposed Adaptive Fusion Mechanism. We plot values of four learnable weights, \emph{i.e.}, $\lambda_{conv}^{cls}$, $\lambda_{fc}^{cls}$, $\lambda_{conv}^{reg}$, and $\lambda_{fc}^{reg}$ against number of training iterations on: (a) PASCAL VOC in the first split, (b) MS COCO, in the base training phase. \textit{cls-conv, cls-fc, reg-conv, reg-fc} indicate $\lambda_{conv}^{cls}$, $\lambda_{fc}^{cls}$, $\lambda_{conv}^{reg}$, and $\lambda_{fc}^{reg}$, respectively.}
	\vspace{-0.3cm}
	\label{fig:6}
\end{figure}

\subsection{Feature Fusion Analysis}\label{E_C}
In this subsection, we discuss the performance of Adaptive Fusion Mechanism. We introduce four learnable weights, \emph{i.e.}, $\lambda_{conv}^{cls}$, $\lambda_{fc}^{cls}$, $\lambda_{conv}^{reg}$, and $\lambda_{fc}^{reg}$ in Eq.~(\ref{eq:3}), to control the contributions of encoded \textit{conv} and \textit{fc} feature components from two paths to the fused output features in Dual Query Encoder and Dual Attention Generator shown in Fig.\ref{fig:3}. Obviously, larger $\lambda_{i}^{j}$ with $i \in\{\textit{conv}, \textit{fc}\}, j \in\{\textit{cls}, \textit{reg}\}$ indicates that extractor $i$ dominates the subtask $j$ in feature extraction.

We visualize the variation of these four weights in base training phase ($K=200$) on PASCAL VOC and MS COCO benchmarks, as shown in Fig.~\ref{fig:6}. As for PASCAL VOC, it is illustrated that the stable value of $\lambda_{conv}^{cls}$ is larger than $\lambda_{fc}^{cls}$ while $\lambda_{conv}^{reg}$ is lower compared with $\lambda_{fc}^{reg}$, representing that \textit{conv} extractor dominates the classification subtask but is less important than \textit{fc} extractor in bounding box regression in this experimental setup and implementation. Besides, we note that $\lambda_{i}^{cls}$ with $i \in\{\textit{conv}, \textit{fc}\}$ in classification subtask is larger than $\lambda_{i}^{reg}$ with $ i \in\{\textit{conv}, \textit{fc}\}$ in regression subtask  respectively, indicating that performing classification generally needs more information than location estimation. When it comes to MS COCO, \textit{conv} extractor consistently leads the classification subtask while these two kinds of extractors almost have equal contributions to bounding box regression. Besides, after stabilizing, all these four weights are larger than those in the experiments on PASCAL VOC respectively, showing that training on complex MS COCO dataset generally requires richer information. Analysis above validates the efficacy of Adaptive Fusion Mechanism when facing distinct subtasks and datasets.

%--------------------------------------------------------------------------
\begin{figure*}[t] \centering
	\vspace{-0.2cm}
	\begin{center}
		\includegraphics[width=1.0\linewidth]{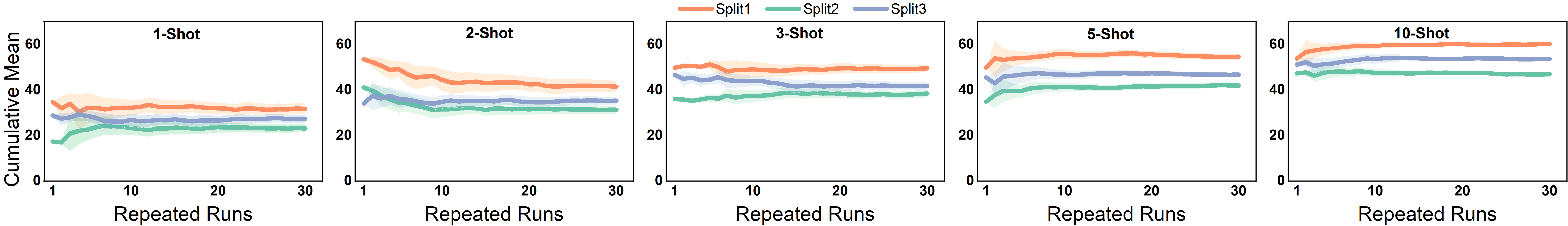}
	\end{center}
	\vspace{-0.2cm}
	\caption{\small Ablation study on effect of multiple runs on PASCAL VOC. We plot cumulative means of $AP_{50}$ with 95\% confidence intervals across 30 repeated runs on novel classes of three novel sets. The means and variances have been stable before 30 runs.}
	\vspace{-0.15cm}
	\label{fig:7}
\end{figure*}

\vspace{-0.1cm}
\subsection{Ablation Study}\label{E_D}

In this subsection, we conduct relative ablations in 10-shot scenario on the first base/novel split of PASCAL VOC. Results in all experiments are obtained by averaging over 10 random runs, as 10 is sufficient to provide statistically stable results in this scenario.

%------------------------------------------------------------------
\begin{table}[] \centering
	\small
	\setlength{\tabcolsep}{1.1mm}
	\caption{\small Ablation study on effect of feature fusion. Comparison results among various fusion combinations in terms of $AP_{50}$ for novel categories on Novel Set 1 of PASCAL VOC dataset are reported.}
	\vspace{-0.2cm}
	{\begin{tabular}{c|cccc|c}
			\hline
			Ablation & cls-conv & cls-fc & reg-conv & reg-fc & Novel $AP_{50}$ \\ \hline%\midrule[0.9pt]
			1        &          & \checkmark      & \checkmark        &        & 49.9       \\
			2        & \checkmark        &        &          & \checkmark      & 56.2       \\
			3        & \checkmark        &        & \checkmark        &        & 55.5       \\
			4        &          & \checkmark      &          & \checkmark      & 55.1       \\
			5        & \checkmark        & \checkmark      &          & \checkmark      & 59.1       \\
			6        & \checkmark        & \checkmark      & \checkmark        &        & 57.6       \\
			7        & \checkmark        &        & \checkmark        & \checkmark      & 59.0       \\
			8        & \checkmark        & \checkmark      & \checkmark        & \checkmark      & \textbf{60.7}       \\ \hline%\bottomrule[1.1pt]
	\end{tabular}}
	\label{tab:4}
	\vspace{-0.2cm}
\end{table}

{\bfseries Effect of Feature Fusion.}
To examine the effectiveness of Adaptive Fusion Mechanism in the design of Dual Query Encoder and Dual Attention Generator, we compare the performance of various feature fusion combinations and report $AP_{50}$ of novel classes in 10-shot setup, as shown in Tabel \ref{tab:4}, where $j\text-i,j \in\{\textit{cls}, \textit{reg}\}, i \in\{\textit{conv}, \textit{fc}\}$ in the header represents applying extractor \textit{i} in subtask \textit{j}. The existence of two extractors within a branch indicates that the information fusion is applied (\emph{e.g.}, the regression branch in Fig.\ref{fig:3} can be represented by ``\textit{reg-fc} \textit{\&} \textit{reg-conv}''). In \textit{Ablation 1-4}, we adopt only one extractor in each branch without feature fusion. Among these four experiments, ``\textit{cls-conv} \textit{\&} \textit{reg-fc}'' achieves the best performance, while ``\textit{cls-fc} \textit{\&} \textit{reg-conv}'' performs the worst, indicating that \textit{conv} extractor prefers classification and \textit{fc} extractor is more suitable for bounding box regression in the first split on PASCAL VOC dataset. This is consistent with the observations in Section \ref{E_C}. In \textit{Ablation 5-7}, feature fusion is adopted in only one branch and boost the detection performance, suggesting that leveraging information from unfocused tasks is essential in each subtask. Applying feature fusion in both branches, as shown in \textit{Ablation 8}, obtains the best performance, further verifying the effectiveness of Adaptive Fusion Mechanism.

\begin{table}[] \centering
	\small
	\vspace{-0.01cm}
	\setlength{\tabcolsep}{1.7mm}
	\caption{\small Ablation study on effect of meta losses. Comparison results among various meta loss combinations in terms of $AP_{50}$ for both base and novel categories on Novel Set 1 of PASCAL VOC dataset are reported.}
	\vspace{-0.2cm}
	\begin{tabular}{c|cc|c|c}
		\hline
		Ablation & meta-cls                  & meta-reg                  & Base $AP_{50}$                    & Novel $AP_{50}$                            \\ \hline
		1        &                           &                           & \textbf{70.8}                & 57.3                                  \\
		2        & \checkmark                         &                           & 70.3                         & 58.9                                  \\
		3        &                           & \checkmark                         & \textbf{70.8}                & 58.4                                  \\
		4        & \checkmark & \checkmark & 70.6 & \textbf{60.7} \\ \hline
	\end{tabular}
	\vspace{-0.3cm}
	\label{tab:5}
\end{table}

{\bfseries Effect of Meta Losses.}
In our approach, we further divide meta loss $L_{meta}$ introduced in~\cite{yan2019meta} into task-specific meta losses, \emph{i.e.}, $L_{meta\text-cls}$  for classification and $L_{meta\text-reg}$ for bounding box regression. Similarly, we evaluate the effect of meta losses in Table \ref{tab:5}. Obviously, both the introduced $L_{meta\text-cls}$ and $L_{meta\text-reg}$ can indeed improve the performance on novel classes. Besides, they provide similar contributions individually. Interestingly, the existence of meta losses almost have no positive influence on the detection of base classes. We believe the reason is that, the few-shot detector has already been able to detect objects from base classes after base training phase with abundant annotations, despite the lack of meta losses.

%--------------------------------------------------------------------------
\begin{figure}[t] \centering
\begin{center}
\includegraphics[width=0.85\linewidth]{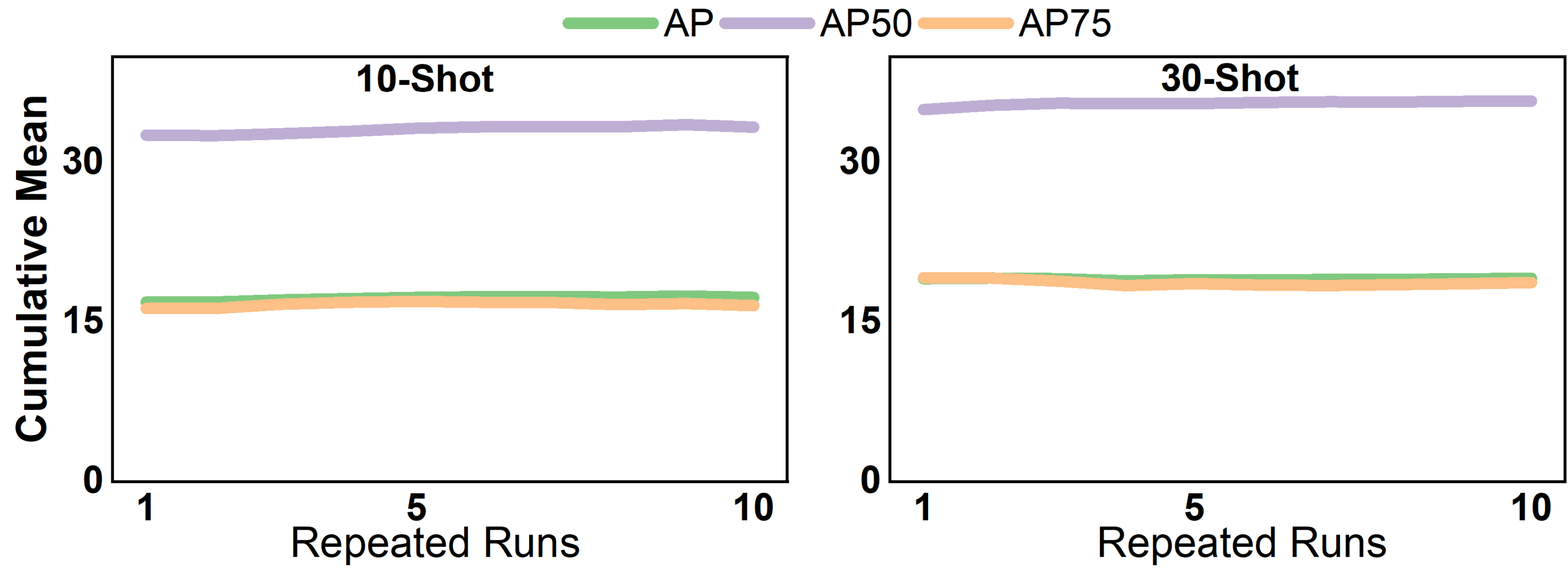}
\end{center}
\vspace{-0.25cm}
\caption{\small Ablation study on effect of multiple runs on MS COCO. We plot cumulative means with 95\% confidence intervals across 10 repeated runs on novel classes. The means are stable and the variances are notably small, consistently.}
\vspace{-0.4cm}
\label{fig:8}
\end{figure}

{\bfseries Effect of Multiple Runs.}
In this ablation study, we focus on the effect of repeated runs in obtaining reported results on PASCAL VOC and MS COCO, as shown in Fig.~\ref{fig:7} and Fig.~\ref{fig:8}, respectively. For PASCAL VOC, the confidence intervals are large across the first few runs (even up to 10 runs, in 3-shot setting), indicating the existence of large sample variance in low-shot scenarios. Therefore, overestimating the actual performance would occur when using only the first few random samples of training shots, as can be observed in the plot of 2-shot case in Fig.~\ref{fig:7}, leading to the unreliability of performance comparisons. In this paper, we adopt 30 repeated runs to obtain stable reported results with smaller variance. For MS COCO, we can observe from Fig.~\ref{fig:8} that, the cumulative means are stable and the confidence intervals are quite small consistently, indicating the robustness of our model. In this paper, we adopt 10 repeated runs for fair comparisons with other methods.

\vspace{-0.2cm}
\section{Conclusion}
\label{sec:Conclusion}
This work targets the problem of few-shot object detection (FSOD). We carefully analyzed the characteristics of FSOD and presented that a general few-shot detector should: 1) explicitly decompose the process of category classification and localization; 2) utilize information from both subtasks for enhancing feature representations. Based on our observations, we proposed Adaptive Fully-Dual Network (AFD-Net) that performs two subtasks separately, involving feature representations, model reweighting, and state estimation. For the acquisition of enhanced features, we proposed a novel Adaptive Fusion Mechanism to guide the design of the feature extractor. Despite its simplicity, our approach achieved significant performance on multiple benchmarks, demonstrating its effectiveness and generalization ability. It is worth noting that, our proposed framework and feature fusion mechanism are general and simple, thus can be potentially applied into other two-stage models, and the performance could be further improved with carefully designed architectures.

\bibliographystyle{ACM-Reference-Format}
\bibliography{egbib}

\end{document}